\documentclass{sig-alternate}
\usepackage{mathrsfs}

\def\equationname{Equation}
\def\sectionname{Section}
\def\algorithmname{Algorithm}
\def\procedurename{Procedure}
\def\appendixname{Appendix}
\def\tablename{Table}
\def\figurename{Figure}
\usepackage[lined,ruled,linesnumbered]{algorithm2e}
\usepackage{subfigure}
\usepackage{multirow}
\usepackage{array}
\usepackage{amsmath}
\usepackage{hyphenat}

\begin{document}
%
\conferenceinfo{CIKM}{'2013 San Francisco Airport Marriott Waterfront, Burlingame, CA, USA.}

\title{Characterizing A Database of Sequential Behaviors with Latent Dirichlet Hidden Markov Models}
\subtitle{[Submitted for Blind Review]
}
%
%
%
%
%

\numberofauthors{1} 
%
\author{
}

\maketitle
\begin{abstract}
This paper proposes a generative model, the latent Dirichlet hidden Markov models (LDHMM), for characterizing a database of sequential behaviors (sequences). LDHMMs posit that each sequence is generated by an underlying\linebreak Markov chain process, which are controlled by the corresponding parameters (i.e., the initial state vector, transition matrix and the emission matrix). These {\it sequence-level} latent parameters for each sequence are modelled as latent Dirichlet random variables and parameterized by a set of deterministic {\it database-level} hyper-parameters. Through this way, we expect to model the sequence in two levels: the database level by deterministic hyper-parameters and the sequence-level by latent parameters. To learn the deterministic hyper-parameters and approximate posteriors of parameters in LDHMMs, we propose an iterative algorithm under the variational EM framework, which consists of E and M steps. We examine two different schemes, the fully-factorized and partially-factorized forms, for the framework, based on different assumptions. We present empirical results of behavior modeling and sequence classification on three real-world data sets, and compare them to other related models. The experimental results prove that the proposed LDHMMs produce better generalization performance in terms of log-likelihood and deliver competitive results on the sequence classification problem.
\end{abstract}





\section{Introduction} \label{Section:Introduction}
In this paper we explore the problem of characterizing a database of sequential behaviors (i.e., sequences). An example of such sequential behaviors is the web browsing behaviors of Internet users. \tablename~\ref{Table:WebBrowsingExample} shows some user-browsing data excerpted from the web server logs of msnbc.com. Each row of the table is an ordered list of discrete symbols, each of which represents one behavior made by a user. Here the behavior is described by the categories of web pages requested by the user. For example, User 1 first browses a `frontpage' page, then visits a `news' page, followed by visiting `travel' page and other pages denoted by dots. This form typical sequential behaviors for a user, and other individuals have similar sequential behaviors. For the last decade, many efforts have been made to characterize the above sequential behaviors for further analysis.

Significant progress has been made on behaviour modelling in the field of Sequence Pattern Mining (SPM). Pattern is an
expression describing a subset of the data \cite{Piateski1991Knowledge}. Sequential pattern mining discovers frequently occurring behaviors or subsequences as patterns, which was first introduced by Agrawal and Srikant \cite{Agrawal1995Mining}. Several algorithms, such as Generalized Sequential Patterns (GSP) \cite{Agrawal1995Mining}, SPADE \cite{Zaki2001SPADE} and PrefixSpan \cite{Han2001PrefixSpan,Han2004Mining}, have been proposed to mine sequential patterns efficiently. Generally speaking, SPM techniques aim at discovering comprehensible sequential patterns in data, which is {\it descriptive} \cite{Novak2009Supervised} and lacks of well-founded theories to apply the discovered patterns for further data analysis tasks, such as sequence classification and behavior modeling.

In the statistical and machine learning community, researchers try to characterize the sequential behaviors using probabilistic models. The probabilistic models not only describe the generative process of the sequential behaviors but also have a {\it predictive} property which is helpful for further analytic tasks, such as prediction of future behaviors. One representative model widely used is the hidden Markov models (HMMs) \cite{Rabiner1990A}. Usually, each sequence is modelled as an HMM. In other words, the dynamics of each sequence is represented by a list of deterministic parameters (i.e., initialization prior vector and transition matrix), and there is no generative probabilistic model for these numbers. This leads to several problems when we directly extend HMMs to modeling a database of sequences: (1) the number of parameters for the HMMs grows linearly with the number of sequences, which leads to a serious problem of over-fitting, and (2) it is not clear how to assign probability to a sequence outside of the training set. Although \cite{Rabiner1990A} suggests a strategy for modeling multiple sequences, it simply ignores the difference on parameters between sequences and assumes all the dynamics of sequences can be characterized by one set of deterministic parameters. This could alleviate the problem of over-fitting to some extent, but may overlook the individual characteristics for individual sequences at the same time, which may further deteriorate the accuracy of behavior modeling.

\begin{table}[!t]
\renewcommand{\arraystretch}{1}
\caption{An Example of Sequential Behaviors}
\label{Table:WebBrowsingExample}
\begin{center}
\begin{tabular}{|c|c c c c|}
  \hline
  User & \multicolumn{4}{|c|}{Sequential Behaviors} \\ \hline \hline
  1    & frontpage & news    & travel    & \ldots   \\ \hline
  2    & news      & news    & news      & \ldots     \\ \hline
  3    & frontpage & news    & frontpage & \ldots     \\ \hline
  4    & frontpage & news    & news      & \ldots   \\ \hline
\end{tabular}
\end{center}
\end{table}

The goal of this paper is to characterize a database of sequential behaviors preserving the essential statistical relationships for each individual sequence and the whole database, while avoiding the problem of over-fitting. To achieve this goal, we propose a generative model that has both sequence-level and database-level variables to comprehensively and effectively modeling behavioral sequences.

The paper is organized as follows: \sectionname~\ref{Section:ProblemStatement} formalizes the problem studied in this paper, followed by the proposed approach described in \sectionname~\ref{Section:LDHMMs}. Then \sectionname~\ref{Section:RelatedModels} reviews the probabilistic models related to this paper. After that, experimental results on several data mining tasks on 3 real-world data sets are reported in \sectionname~\ref{Section:EmpiricalStudy}. Finally, \sectionname~\ref{Section:Conclusions} concludes this paper and discusses about some possible future directions.

\section{Problem Statement} \label{Section:ProblemStatement}
Here we use the terminologies, such as ``behaviors'', ``sequences'' and ``database'', to describe a database of sequential behaviors. This is helpful for understanding the probabilistic model derived on the data. It is important to note that the model proposed in this paper is also applicable to other sequential data that has the similar data forms. In this paper, vectors are denoted by lower case bold Roman or Greek letters and all vectors are assumed to be column vectors except for special explanations. Uppercase bold Roman letters denote matrices while letters in other cases are assumed to be scalar.

\begin{itemize}
\item A database $\mathscr{D}$ is a collection of $M$ sequences denoted by $\mathscr{D} = \{\mathbf{X}_{1}, \mathbf{X}_{2}, \cdots, \mathbf{X}_{M} \}$.

\item A sequence $\mathbf{X}_{m}$ ($1 \le m \le M$) is an ordered list of $N_{m}$ behaviors denoted by $\mathbf{X}_{m} = (\mathbf{x}_{m 1}, \mathbf{x}_{m 2}, \cdots, \mathbf{x}_{m N_{m}})$, where $\mathbf{x}_{m n}$ ($1 \le n \le N_{m}$) is the $n^{th}$ behavior in the sequence $\mathbf{X}_{m}$. The behaviors in the sequence are ordered by increasing time when behaviors are made.

\item A behavior $\mathbf{x}_{m n}$ ($1 \le m \le M$, $1 \le n \le N_{m}$) is the basic unit of sequential behaviors, defined to be a 1-of-$V$ vector $\mathbf{x}_{m n}$ such that $x_{m n v} = 1$ and $x_{m n u} = 0$ (for all $u \not= v$), which represents an item $v$ from a vocabulary indexed by $\{ 1, 2, \cdots, V \}$. Each index represents one type of behaviors, such as browsing a `travel' web page as shown in \tablename~\ref{Table:WebBrowsingExample}.

\end{itemize}

Given a database $D$ of sequential behaviors, the problem of characterizing behaviors  is to derive a probabilistic model which preserves the statistical relationships in the sequences and tends to assign high likelihood to ``similar'' sequences.

\section{The Proposed Model} \label{Section:LDHMMs}
\subsection{The Graphical Model}
The basic idea of the Latent Dirichlet Hidden Markov Models (LDHMMs) is that the dynamics of each sequence $\mathbf{X}_{m}$ is assumed to be reflected through a hidden Markov chain $\mathbf{Z}_{m} = (\mathbf{z}_{m 1}, \mathbf{z}_{m 2}, \cdots, \mathbf{z}_{m N_{m}})$\footnote{$\mathbf{z}_{m n}$ ($1 \le n \le N_{m}$) can be represented by a 1-of-$K$ vector (similar to the form of a behavior $\mathbf{x}_{mn}$) and has $K$ possible hidden states, where $K$ is the number of possible hidden states and is usually set empirically.} parameterized by the corresponding initial prior vector $\boldsymbol{\pi}_{m}$, transition matrix $\mathbf{A}_{m}$ and a state-dependent emission matrix $\mathbf{B}_{m}$. Then $\boldsymbol{\pi}_{m}$, $\mathbf{A}_{m}$ ($\mathbf{a}_{m i}$ ($1 \le i \le K$) is the $i$th row vector of $\mathbf{A}_{m}$) and $\mathbf{B}_{m}$ ($\mathbf{b}_{m i}$ ($1 \le i \le K$) is the $i$th row vector of $\mathbf{B}_{m}$) can be seen as a lower dimension representation of the dynamics of the sequence. The distribution of these parameters of all sequences are then further governed by database-level Dirichlet hyper-parameters, i.e., $\boldsymbol{\alpha}^{(\pi)}$, $\boldsymbol{\alpha}^{(A)}_{1:K}$ and $\boldsymbol{\beta}_{1:K}$, where $\boldsymbol{\alpha}^{(A)}_{1:K}$ is a matrix whose $i$th row vector is $\boldsymbol{\alpha}^{(A)}_{i}$ and $\boldsymbol{\beta}_{1:K}$ is a matrix whose $i$th row vector is $\boldsymbol{\beta}_{i}$. To be more specific, for a database of sequential behaviors $\mathscr{D}$, the generative process is as follows\footnote{Please refer to \appendixname~\ref{Appendix:Distributions} for details of Dirichlet (Dir) and Multinomial distributions.}:
\begin{enumerate}
\item[1.] Generate hyper-parameters $\boldsymbol{\alpha}^{(\pi)}$, $\boldsymbol{\alpha}^{(A)}_{1:K}$ and $\boldsymbol{\beta}_{1:K}$.
\item[2.] For each sequence index $m$,
  \begin{enumerate}
  \item[1)] Generate $\boldsymbol{\pi}_{m} \sim Dir(\boldsymbol{\alpha}^{(\pi)})$, $\mathbf{a}_{m i} \sim Dir(\boldsymbol{\alpha}^{(A)}_{i})$ and $\mathbf{b}_{m i} \sim Dir(\boldsymbol{\beta}_{i})$
  \item[2)] For the first time stamp in the sequence $\mathbf{X}_{m}$:
      \begin{enumerate}
      \item[(a)] Generate an initial hidden state \linebreak $\mathbf{z}_{m1} \sim Multinomial(\pi)$.
      \item[(b)] Generate a behavior from $p(\mathbf{x}_{m1} | \mathbf{z}_{m1}, \mathbf{B}_{m})$, a multinomial probability conditioned on the hidden state $\mathbf{z}_{m1}$ and $\mathbf{B}_{m}$.
      \end{enumerate}
  \item[3)] For each of other time stamps in the sequence $\mathbf{X}_{m}$ ($1 \le n \le N_{m}$):
      \begin{enumerate}
      \item[(a)] Generate a hidden state $\mathbf{z}_{m n}$ from \linebreak $p(\mathbf{z}_{m n} | \mathbf{z}_{m, n-1}, \mathbf{A_{m}})$.
      \item[(b)] Generate a behavior from $p(\mathbf{x}_{m n} | \mathbf{z}_{m n}, \mathbf{B}_{m})$.
      \end{enumerate}
  \end{enumerate}
\end{enumerate}

\begin{figure}[!t]
\begin{center}
\includegraphics[scale = 0.5]{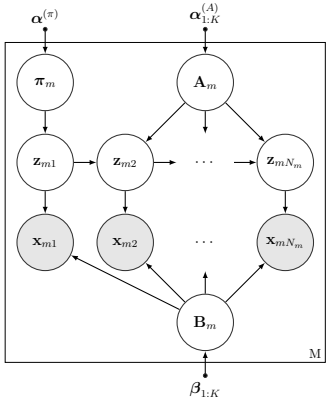}
\end{center}
\caption{The Graphical Model for the LDHMMs.}
\label{Figure:LDHMMs}
\end{figure}

Accordingly, the graphical model of LDHMMs is shown in \figurename~\ref{Figure:LDHMMs}. As per the graph states itself, there are three levels of modeling in LDHMMs. The hyper-parameters $\boldsymbol{\alpha}^{(\pi)}$, $\boldsymbol{\alpha}^{(A)}_{1:K}$ and $\boldsymbol{\beta}_{1:K}$ are database-level variables, assumed to be sampled once in the process of generating a database. The variables $\boldsymbol{\pi}_{m}$, $\mathbf{A}_{m}$ and $\mathbf{B}_{m}$ ($1 \le m \le M$) are sequence-level variables, denoted as $\boldsymbol{\theta}_{m} = \{\boldsymbol{\pi}_{m}, \mathbf{A}_{m}, \mathbf{B}_{m}\}$ sampled once per sequence. Finally, the variables $\mathbf{z}_{m n}$ and $\mathbf{x}_{m n}$ are behavior-level variables sampled once for each behavior in each sequence.

\subsection{Learning the Model}
In this section, our goal is to learn the deterministic hyper-parameters of the LDHMMs given a database $D$, which is to maximize the likelihood function $\log p(\mathscr{D}; \boldsymbol{\alpha}^{(\pi)}, \boldsymbol{\alpha}^{(A)}_{1:K}, \boldsymbol{\beta}_{1:K})$ given by
\begin{equation}
   \sum_{\mathbf{Z}_{m}, m} \int_{\boldsymbol{\theta}_{m}} \log p(\mathbf{X}_{m}, \mathbf{Z}_{m}, \boldsymbol{\theta}_{m}; \boldsymbol{\alpha}^{(\pi)}, \boldsymbol{\alpha}^{(A)}_{1:K}, \boldsymbol{\beta}_{1:K})
\end{equation}

Direct optimization of the above equation is very difficult since the involvement of latent variables, thus we turn to optimize its lower bound $L(q, \boldsymbol{\alpha}^{(\pi)}, \boldsymbol{\alpha}^{(A)}_{1:M}, \boldsymbol{\beta}_{1:K})$ given by the Jensen's inequality as follows \cite{Bishop2006Pattern}:
\begin{equation} \label{Equation:LowerBound}
    L = \sum_{m = 1}^{M} [ E_{q} [ \log p(\boldsymbol{\theta}_{m}, \mathbf{Z}_{m}, \mathbf{X}_{m}) ] - E_{q} [ \log q(\boldsymbol{\theta}_{m}, \mathbf{Z}_{m}) ] ]
\end{equation}
where $q$ is assumed to be variational distribution function approximating to the posterior distribution of latent variables $\boldsymbol{\theta}_{m}$ given $\mathbf{X}_{m}, \boldsymbol{\alpha}^{(\pi)}, \boldsymbol{\alpha}^{(A)}_{1:K}, \boldsymbol{\beta}_{1:K}$ and can be decomposed as $q_{1} q_{2} \cdots q_{M}$. Specifically, $q_{m} = q_{m} (\boldsymbol{\theta}_{m}, \mathbf{Z}_{m})$ is a variational distribution function approximating to \linebreak $p ( \mathbf{\theta}_{m}, \mathbf{Z}_{m} | \mathbf{X}_{m}; \boldsymbol{\alpha}^{(\pi)}, \boldsymbol{\alpha}^{(A)}_{1:K}, \boldsymbol{\beta}_{1:K}) $ for the sequence $\mathbf{X}_{m}$.

Then the lower bound of the likelihood function becomes a function of $q$, $\boldsymbol{\alpha}^{(\pi)}$, $\boldsymbol{\alpha}^{(A)}_{1:K}$ and $\boldsymbol{\beta}_{1:K}$. To obtain the optimal $\boldsymbol{\alpha}^{(\pi)*}$, $\boldsymbol{\alpha}^{(A)*}_{1:K}$ and $\boldsymbol{\beta}^{*}_{1:K}$ is still difficult since the involvement of $q$. Thus, we propose a variational EM-based algorithm for learning the hyper-parameters of the LDHMMs, which yields the algorithm summarized in \algorithmname~\ref{Algorithm:LDHMMs} and is guaranteed to increase likelihood after each iteration \cite{Bishop2006Pattern}. To be more specific, the variational EM algorithm is a two-stage iterative optimization technique which iterates the E-step (i.e., optimization with respect to $q$) and M-step (optimization with respect to the hyper-parameters) from lines 1 to 7. For each iteration, the E-step (lines 1-4) fixes the hyper-parameters and optimize the $L$ with respect to $q_{m}$ for each sequence; while the M-step (line 5) fixes the $q$ and optimizes the $L$ with respect to the hyper-parameters. Through this manner, the optimal hyper-parameters $\boldsymbol{\alpha}^{(\pi)*}$, $\boldsymbol{\alpha}^{(A)*}_{1:K}$ and $\boldsymbol{\beta}^{*}_{1:K}$ are obtained when the iterations are converged in line 7. It is also important to note that the approximate posteriors of sentence-level parameters (i.e., $q$) are learned as by-products in E steps.

The following two sections will discuss the details of the procedure E-step and M-step in \algorithmname~\ref{Algorithm:LDHMMs} and gives out two different implementations.

\begin{algorithm}[!t]
\SetKwInOut{In}{Input}\SetKwInOut{Out}{Output}
\SetKwFunction{Estep}{Estep}
\SetKwFunction{Mstep}{Mstep}
\In{An initial setting for the parameters $\boldsymbol{\alpha}^{(\pi)}$, $\boldsymbol{\alpha}^{(A)}_{1:K}$, $\boldsymbol{\beta}_{1:K}$}
\Out{Learned hyper-parameters $\boldsymbol{\alpha}^{(\pi)*}$, $\boldsymbol{\alpha}^{(A)*}_{1:K}$, $\boldsymbol{\beta}^{*}_{1:K}$}
\BlankLine
\While{the convergence criterion is not satisfied}{
\tcp{E-step}
\ForEach{sequence $\mathbf{X}_{m}$}{
   \tcp{optimize $L$ with respect to $q_{m}$}
   $q_{m}$ $\leftarrow$ \Estep{$\boldsymbol{\alpha}^{(\pi)}, \boldsymbol{\alpha}^{(A)}_{1:K}, \boldsymbol{\beta}_{1:K}, \mathbf{X}_{m}$} \;
}
\tcp{M-step}
   \tcp{optimizing $L$ with respect to $\boldsymbol{\alpha}^{(\pi)}$, $\boldsymbol{\alpha}^{(A)}_{1:K}$, $\boldsymbol{\beta}_{1:K}$}
   $\boldsymbol{\alpha}^{(\pi)}, \boldsymbol{\alpha}^{(A)}_{1:K}, \boldsymbol{\beta}_{1:K}$ $\leftarrow$ \Mstep{$q$, $\boldsymbol{\alpha}^{(\pi)}, \boldsymbol{\alpha}^{(A)}_{1:K}, \boldsymbol{\beta}_{1:K}$} \;
}
$\boldsymbol{\alpha}^{(\pi)*}, \boldsymbol{\alpha}^{(A)*}_{1:K}, \boldsymbol{\beta}^{*}_{1:K}$ $\leftarrow$ $\boldsymbol{\alpha}^{(\pi)}_{1:K}, \boldsymbol{\alpha}^{(A)}_{1:K}, \boldsymbol{\beta}_{1:K}$ \;
\caption{The Learning Algorithm for LDHMMs.}\label{Algorithm:LDHMMs}
\end{algorithm}

\subsection{The E step: Variational Inference of Latent Variables}
In this section, we provide the details of the E step, which is to estimate $q_{m}$ for ($1 \le m \le M$) given the observed sequence $\mathbf{X}_{m}$ and fixed hyper-parameters $\boldsymbol{\alpha}^{(\pi)}_{1:K}, \boldsymbol{\alpha}^{(A)}_{1:K}, \boldsymbol{\beta}_{1:K}$ and this process is usually termed as variational inference \cite{Bishop2006Pattern,Ghahramani2000VariationalInference,Jaakkola1997Variational,Jordan1999An}.

Here we consider two different implementations of variational inference based on different decompositions of $q_{m}$:
\begin{itemize}
\item A fully-factorized (FF) form.
\item A partially-factorized (PF) form.
\end{itemize}

\begin{figure}[!t]
\begin{center}
\subfigure[]{\label{Figure:LDHMM-ff}{\includegraphics[scale = 0.4]{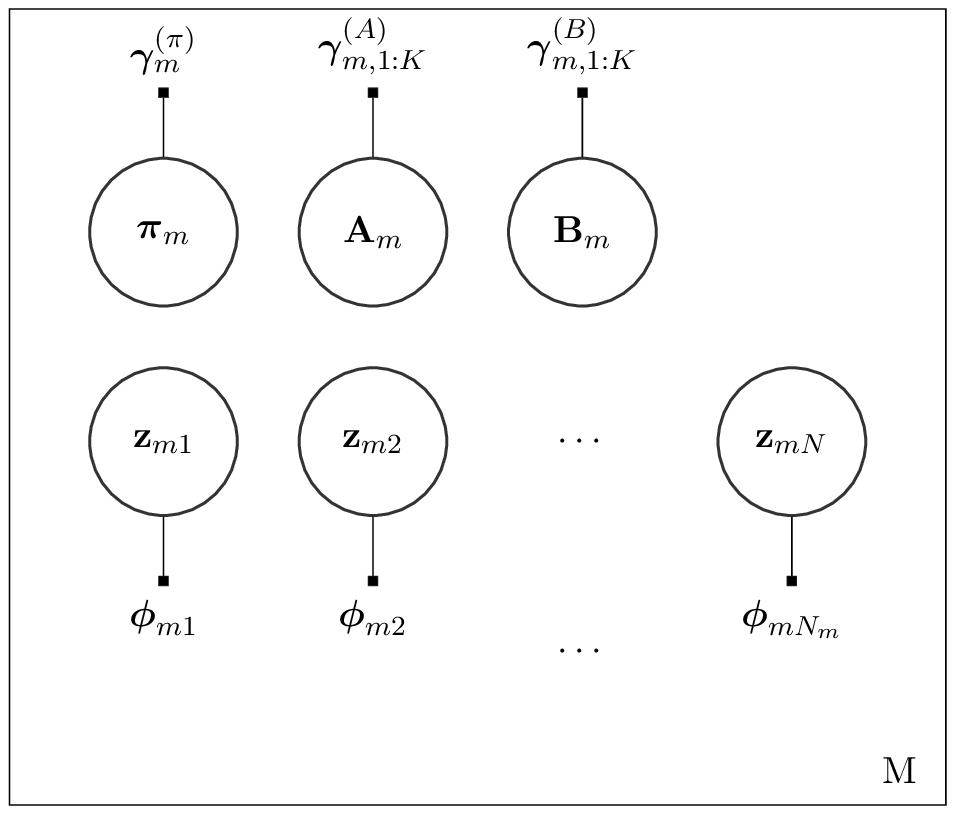}}}
\subfigure[]{\label{Figure:LDHMM-pf}{\includegraphics[scale = 0.4]{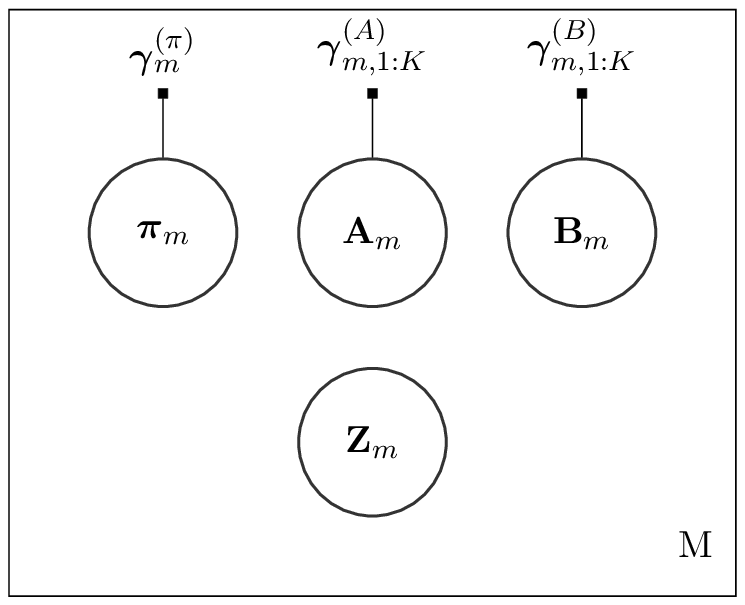}}}
\end{center}
\caption{The graphical models of variational distributions: (a) the FF form, (b)  the PF form.}
\label{Figure:RelatedModels}
\end{figure}

As shown in \figurename~\ref{Figure:LDHMM-ff}, the FF form assumes:
\begin{equation} \begin{aligned}
  q_{m} (\mathbf{\theta}_{m}) &=& q_{m} (\boldsymbol{\pi}_{m}) q_{m} (\mathbf{A}_{m}) q_{m} (\mathbf{B}_{m}) \nonumber \\
                              & &  q_{m}(\mathbf{z}_{m1}) q_{m}(\mathbf{z}_{m2}) \cdots q_{m}(\mathbf{z}_{mN_{m}})
\end{aligned} \end{equation}
This is inspired by the standard mean-field approximation in \cite{Jaakkola1997Variational,Jordan1999An}.

As shown in \figurename~\ref{Figure:LDHMM-pf}, the PF form assumes:
\begin{equation}
  q_{m} (\mathbf{\theta}_{m}) = q_{m} (\boldsymbol{\pi}_{m}) q_{m} (\mathbf{A}_{m}) q_{m} (\mathbf{B}_{m}) q_{m}(\mathbf{Z}_{m})
\end{equation}
and no further assumption has been made on $q_{m}(\mathbf{Z}_{m})$. This is inspired by the manners proposed in \cite{Mckay1997Ensemble,Ghahramani1997On,Beal2003Variational,Ghahramani2000VariationalLearning}, which preserves the conditional dependency between the variables of $q_{m}(\mathbf{Z}_{m})$.

\subsubsection{E Step: the FF Form}
The variational inference process of the FF form yields the following iterations. Please refer to \appendixname~\ref{Appendix:VI-FF} for the details of the derivation of the updating formulas.

\paragraph{Fixed $\boldsymbol{\gamma}^{(\pi)}_{m}$ and $\boldsymbol{\gamma}^{(B)}_{m, 1:K}$ and $\boldsymbol{\gamma}^{(A)}_{m, 1:K}$, Update $\boldsymbol{\phi}_{m, 1:N_{m}}$ ($1 \le m \le M$)}
$\phi_{m 1 i}$ is updated by:
\begin{equation} \begin{array}{l} \label{Equation:Update-z1-ff}
  \phi_{m 1 i} = exp [\sum_{j = 1}^{V} (x_{m 1 j} ( \Psi (\gamma^{(B)}_{m i j}) - \Psi (\sum_{v = 1}^{V} \gamma^{(B)}_{m i v})))  \\
               +  (\Psi (\gamma^{(\pi)}_{m i}) - \Psi (\sum_{j = 1}^{K} \gamma^{(\pi)}_{m j})) \\
               +  \sum_{k = 1}^{K} \phi_{m 2 k} (\Psi (\gamma^{(A)}_{m i k}) - \Psi (\sum_{j = 1}^{K} \gamma^{(A)}_{m i j}))]
\end{array} \end{equation}

$\phi_{m n i}$ ($2 \le n \le N_{m}-1$) is updated by:
\begin{equation} \begin{array}{l} \label{Equation:Update-z2toNminus1-ff}
  \phi_{m n i} = exp [\sum_{j = 1}^{V} (x_{m n j} ( \Psi (\gamma^{(B)}_{m i j}) - \Psi (\sum_{v = 1}^{V} \gamma^{(B)}_{m i v})))  \\
               +  (\Psi (\gamma^{(\pi)}_{m i}) - \Psi (\sum_{j = 1}^{K} \gamma^{(\pi)}_{m j}))  \\
               +  \sum_{i = 1}^{K} \phi_{m, n-1, i} (\Psi (\gamma^{(A)}_{m i k}) - \Psi (\sum_{j = 1}^{K} \gamma^{(A)}_{m i j}))]  \\
               +  \sum_{k = 1}^{K} \phi_{m, n+1, k} (\Psi (\gamma^{(A)}_{m i k}) - \Psi (\sum_{j = 1}^{K} \gamma^{(A)}_{m i j}))]
\end{array} \end{equation}

$\phi_{m N_{m} i}$ is updated by:
\begin{equation} \begin{array}{l} \label{Equation:Update-zN-ff}
  \phi_{m N_{m} i} =  exp [\sum_{j = 1}^{V} (x_{m N_{m} j} ( \Psi (\gamma^{(B)}_{m i j}) \\ - \Psi (\sum_{v = 1}^{V} \gamma^{(B)}_{m i v})))
                   +  (\Psi (\gamma^{(\pi)}_{m i}) - \Psi (\sum_{j = 1}^{K} \gamma^{(\pi)}_{m j}))  \\
                   +  \sum_{i = 1}^{K} \phi_{m, N_{m-1}, i} (\Psi (\gamma^{(A)}_{m i k}) - \Psi (\sum_{j = 1}^{K} \gamma^{(A)}_{m i j}))]
\end{array} \end{equation}

\paragraph{Fixed $\boldsymbol{\phi}_{m, 1:N_{m}}$, $\boldsymbol{\gamma}^{(A)}_{m, 1:K}$, $\boldsymbol{\gamma}^{(B)}_{m, 1:K}$, Update $\boldsymbol{\gamma}^{(\pi)}_{m}$}
\begin{equation} \label{Equation:Update-pi-ff}
   \gamma^{(\pi)}_{m i} = \alpha^{(\pi)}_{i} + \phi_{m 1 i}
\end{equation}

\paragraph{Fixed $\boldsymbol{\phi}_{m, 1:N_{m}}$, $\boldsymbol{\gamma}^{(\pi)}_{m}$ and $\boldsymbol{\gamma}^{(B)}_{m, 1:K}$, Update $\boldsymbol{\gamma}^{(A)}_{m, 1:K}$}
\begin{equation} \label{Equation:Update-A-ff}
   \gamma^{(A)}_{m i k} = \alpha^{(A)}_{i k} + \sum_{n = 2}^{N} \phi_{m, n-1, i} \phi_{m, n, k}
\end{equation}

\paragraph{Fixed $\boldsymbol{\phi}_{m, 1:N_{m}}$, $\boldsymbol{\gamma}^{(\pi)}_{m}$ and $\boldsymbol{\gamma}^{(A)}_{m, 1:K}$, Update $\boldsymbol{\gamma}^{(B)}_{m, 1:K}$}
\begin{equation} \label{Equation:Update-B-ff}
   \gamma^{(B)}_{m i j} = \alpha^{(A)}_{i j} + \sum_{n = 1}^{N} \phi_{m n i} x_{m n j}
\end{equation}

For simplicity, the E step for the FF form can be summarized in \procedurename~\ref{Procedure:Estep-ff}.

\begin{procedure}[!t]
\SetKwInOut{In}{input}\SetKwInOut{Out}{output}
\In{A set of the parameters $\boldsymbol{\alpha}^{(\pi)}, \boldsymbol{\alpha}^{(A)}_{1:K}, \boldsymbol{\beta}_{1:K}$, a sequence $\mathbf{X}_{m}$}
\Out{The variational distribution $q_{m}$}
\BlankLine
Initialize $\gamma^{(\pi)}_{m i}$ for all $i$ \;
Initialize $\gamma^{(A)}_{m i k}$ for all $i$ and $k$ \;
Initialize $\gamma^{(B)}_{m i j}$ for all $i$ and $j$ \;

\Repeat{convergence}{
  Update $\phi_{m n i}$ according to \equationname~\ref{Equation:Update-z1-ff} to \ref{Equation:Update-zN-ff} for all $n$ and $i$\;
  Update $\gamma^{(\pi)}_{m i}$ according to \equationname~\ref{Equation:Update-pi-ff} for all $i$\;
  Update $\gamma^{(A)}_{m i k}$ according to \equationname~\ref{Equation:Update-A-ff} for all $i$ and $k$\;
  Update $\gamma^{(B)}_{m i j}$ according to \equationname~\ref{Equation:Update-B-ff} for all $i$ and $j$\;
}

$q_{m}$ $\leftarrow$ $q_{m}(Z_{m})$, $q_{m}(\boldsymbol{\pi}_{m})$, $q_{m}(\mathbf{A}_{m})$, $q_{m}(\mathbf{B}_{m})$ \;
\caption{Estep($\boldsymbol{\alpha}^{(\pi)}, \boldsymbol{\alpha}^{(A)}_{1:K}, \boldsymbol{\beta}_{1:K}, \mathbf{X}_{m}$)}\label{Procedure:Estep-ff}
\end{procedure}

\subsubsection{E Step: the PF Form} \label{Subsubsection:PF}
The variational inference process of the PF form yields the following iterations. Please refer to \appendixname~\ref{Appendix:VI-PF} for the details of the derivation of the updating formulas.

\paragraph{Fixed $\boldsymbol{\gamma}^{(\pi)}_{m}$ and $\boldsymbol{\gamma}^{(B)}_{m, 1:K}$ and $\boldsymbol{\gamma}^{(A)}_{m, 1:K}$, Update $q_{m}(\mathbf{Z}_{m})$}
The relevant margining posteriors of the distribution $q_{m}(\mathbf{Z}_{m})$, i.e., $q_{m}(\mathbf{z}_{m n}) = \boldsymbol{\gamma}_{m n}$ ($1 \le n \le N_{m}$) and $q_{m}(\mathbf{z}_{m n}, \mathbf{z}_{m,n+1}) = \boldsymbol{\xi}_{m,n, n+1}$ ($1 \le n \le N_{m} - 1$) using the forward-backward algorithm \cite{Rabiner1990A} and the details are described in \appendixname~\ref{Appendix:VI-PF-ForwardBackward}.

\paragraph{Fixed $q_{m}(\mathbf{Z}_{m})$, $\boldsymbol{\gamma}^{(A)}_{m, 1:K}$, $\boldsymbol{\gamma}^{(B)}_{m, 1:K}$, Update $\boldsymbol{\gamma}^{(\pi)}_{m}$}
\begin{equation} \label{Equation:Update-pi-pf}
   \gamma^{(\pi)}_{m i} = \alpha^{(\pi)}_{i} + q_{m}(z_{m 1 i})
\end{equation}
where $1 \le m \le M$ and $1 \le i \le K$.

\paragraph{Fixed $q_{m}(\mathbf{Z}_{m})$, $\boldsymbol{\gamma}^{(\pi)}_{m}$ and $\boldsymbol{\gamma}^{(B)}_{m, 1:K}$, Update $\boldsymbol{\gamma}^{(A)}_{m, 1:K}$}
\begin{equation} \label{Equation:Update-A-pf}
   \gamma^{(A)}_{m i k} = \alpha^{(A)}_{i k} + \sum_{n = 2}^{N} q(z_{m, n-1, i}, z_{m n k})
\end{equation}

\paragraph{Fixed $q_{m}(\mathbf{Z}_{m})$, $\boldsymbol{\gamma}^{(\pi)}_{m}$ and $\boldsymbol{\gamma}^{(A)}_{m, 1:K}$, Update $\boldsymbol{\gamma}^{(B)}_{m, 1:K}$}
\begin{equation} \label{Equation:Update-B-pf}
   \gamma^{(B)}_{m i j} = \alpha^{(B)}_{i j} + \sum_{n = 1}^{N} x_{m n j} q_{m}(z_{m n i})
\end{equation}
where $1 \le m \le M$, $1 \le i \le K$ and $1 \le j \le V$.

The E-step can be also summarized as a procedure similar to \procedurename~\ref{Procedure:Estep-ff} by replacing the corresponding updating formulas. We omit it here for conciseness.

%
%

\subsubsection{Discussion of computational complexity} \label{Subsubsection:Complexity}
The computational complexity for the E-step of approximately inferring the posterior distribution of $\boldsymbol{\pi}_{m}$, $\boldsymbol{B}_{m, 1:K}$ and $\boldsymbol{A}_{m, 1:K}$ ($1 \le m \le M$) given the hyper-parameters and the observed behaviors are similar for both the PF and FF forms. Specifically, the computational complexity for inferring the approximate posteriors of $\boldsymbol{\pi}_{m}$, $\boldsymbol{B}_{m, 1:K}$ and $\boldsymbol{A}_{m, 1:K}$ ($1 \le m \le M$) are the same for the two forms, which are proportional to $ O(M T_{E} K)$, $ O(M T_{E} K^{2} N)$ and $ O(M T_{E} K^{2} N)$, respectively, where $K$ is the number of hidden states, $T_{E}$ is the iteration number of E-step, N is the maximum length of all sequences. However, the computational cost for approximate inference of the posterior of $\mathbf{Z}_{m}$ ($1 \le m \le M$) is slightly different for the two forms. The computational complexity for the PF form is proportional to $O(K^{2}N)$ while its counterpart of the FF form is proportional to $O(KN)$. Thus, the overall computational complexity for the PF and FF form are $O(M T_{E} (K + 3 K^{2} N))$ and $O(M T_{E} (K + K N + 2 K^{2} N))$, respectively. It is clear that two forms have comparable computational cost and the FF form is slightly faster.

\subsection{The M Step: Fixed Variational Variables, Estimate Hyper-parameters}
In this section, we provide the details of the M-step, which is to estimate hyper-parameters $\boldsymbol{\alpha}^{(\pi)}_{1:K}, \boldsymbol{\alpha}^{(A)}_{1:K}, \boldsymbol{\beta}_{1:K}$ given the observed sequence $\mathbf{X}_{m}$ and fixed variational variables $q_{m}$ for ($1 \le m \le M$). In particular, it maximizes the lower bound of the log-likelihood $L$ with respect to respective hyper-parameters as follows:

\subsubsection{The FF Form}
\paragraph{Update $\boldsymbol{\alpha}^{(\pi)}$}
Maximizing $\boldsymbol{\alpha}^{(\pi)}$ can be solved by iterative linear-time Newton-Raphson algorithm \cite{Blei2003Latent,Minka2000Estimating}. Define the following variables:
\begin{equation}
   g_{i} = M (\Psi(\sum_{j = 1}^{K} \alpha^{(\pi)}_{j}) - \Psi(\alpha^{(\pi)}_{i})) + \sum_{m = 1}^{M} (\Psi (\gamma^{(\pi)}_{m i}) - \Psi (\sum_{j = 1}^{K} \gamma^{(\pi)}_{m j})) \label{Equation:Newton-Raphson1-g-pf}
\end{equation}
\begin{equation}
   h_{i} = -M \Psi^{'}(\alpha^{(\pi)}_{i}) \label{Equation:Newton-Raphson1-h-pf}
\end{equation}
\begin{equation}
   w = M \Psi(\sum_{j = 1}^{K} \alpha^{(\pi)}_{j}) \label{Equation:Newton-Raphson1-w-pf}
\end{equation}
\begin{equation}
   c = \frac{\sum_{j = 1}^{K} g_{j} / h_{j}}{{w}^{-1} + \sum_{j = 1}^{K} h_{j}^{-1}} \label{Equation:Newton-Raphson1-c-pf}
\end{equation}

The updating equation is given by:
\begin{equation} \label{Equation:Update-alphapi-pf}
    \alpha^{(\pi)*}_{i} = \alpha^{(\pi)}_{i} - \eta \frac{g_{i} - c}{h_{i}}
\end{equation}

\procedurename~\ref{Procedure:NewtonRaphson1} summarizes the above algorithm, which is an iterative process of updating the value of $\boldsymbol{\alpha}^{(\pi)}$. To be more specific, at the beginning of each iteration, the variables $g, h, w, c$ are calculated by \equationname~\ref{Equation:Newton-Raphson1-g-pf}-\ref{Equation:Newton-Raphson1-c-pf} and $\eta$ to be 1 in lines 2 and 3. Then line 4 updates $\boldsymbol{\alpha}^{(\pi)*}$ by \equationname~\ref{Equation:Update-alphapi-pf} and line 5 judges if the updated $\boldsymbol{\alpha}^{(\pi)*}$ falls into the feasible range. If so, it reduces $\eta$ by a factor of 0.5 in line 6 and updates $\boldsymbol{\alpha}^{(\pi)*}$ in line 7 until it becomes valid. In line 8, update $\boldsymbol{\alpha}^{(\pi)}$ as $\boldsymbol{\alpha}^{(\pi)*}$ for the next iteration.

\begin{procedure}[!t]
\SetKwInOut{In}{input}\SetKwInOut{Out}{output}
\In{ $\gamma_{1:M, 1:K}^{(\pi)}$, $\boldsymbol{\alpha}^{(\pi)}$, Number of iterations $T_{M}$}
\Out{Updated $\alpha^{(\pi)*}_{i}$}
\BlankLine

\For{$iter \leftarrow 1$ \KwTo $T_{M}$}{
   Update $g_{i}, h_{i}, w, c$ for all $i$ according to \equationname~\ref{Equation:Newton-Raphson1-g-pf}-\ref{Equation:Newton-Raphson1-c-pf} \;
   $\eta$ $\leftarrow$ 1\;
   Update $\alpha^{(\pi)*}_{i}$ according to \equationname~\ref{Equation:Update-alphapi-pf}\;
   \While{Any $\alpha^{(\pi)*}_{i} < 0$}{
     $\eta$ $\leftarrow$ $0.5 \eta$\;
     Update $\alpha^{(\pi)*}_{i}$ according to \equationname~\ref{Equation:Update-alphapi-pf}\;
   }
   $\alpha^{(\pi)}_{i}$ $\leftarrow$ $\alpha^{(\pi)*}_{i}$\;
}

\caption{Newton-Raphson($\gamma_{1:M, 1:K}^{(\pi)}, \boldsymbol{\alpha}^{(\pi)}, T_{M}$)} \label{Procedure:NewtonRaphson1}
\end{procedure}

\paragraph{Update $\boldsymbol{\alpha}^{(A)}_{1:K}$}
Similarly, the estimation of $\boldsymbol{\alpha}^{(A)}_{i}$ ($1 \le i \le K$) can be solved by the \procedurename~\ref{Procedure:NewtonRaphson1} with changes on \equationname~\ref{Equation:Newton-Raphson1-g-pf}-\ref{Equation:Update-alphapi-pf} (i.e., replace $\boldsymbol{\alpha}^{(\pi)}$ by $\boldsymbol{\alpha}^{(A)}_{i}$ and $\boldsymbol{\gamma}^{(\pi)}$ by $\boldsymbol{\gamma}^{(A)}_{i}$).

\paragraph{Update $\boldsymbol{\beta}_{1:K}$}
Similarly, the estimation of $\boldsymbol{\beta}_{i}$ ($1 \le i \le K$) can be done by the \procedurename~\ref{Procedure:NewtonRaphson1} with changes on \equationname~\ref{Equation:Newton-Raphson1-g-pf}-\ref{Equation:Update-alphapi-pf} (i.e., replace $\boldsymbol{\alpha}^{(\pi)}$ by $\boldsymbol{\beta}_{i}$ and $\boldsymbol{\gamma}^{(\pi)}$ by $\boldsymbol{\gamma}^{(B)}_{i}$).

The M-step can be summarized in \procedurename~\ref{Procedure:Mstep}.

\begin{procedure}[!t]
\SetKwInOut{In}{input}\SetKwInOut{Out}{output}
\In{$q, \boldsymbol{\alpha}^{(\pi)}, \boldsymbol{\alpha}^{(A)}_{1:K}, \boldsymbol{\beta}_{1:K}$}
\Out{A set of the parameters $\boldsymbol{\alpha}^{(\pi)}, \boldsymbol{\alpha}^{(A)}_{1:K}, \boldsymbol{\beta}_{1:K}$}
\BlankLine

  Call \procedurename~\ref{Procedure:NewtonRaphson1} to update $\boldsymbol{\alpha}^{(\pi)}$ \;
  Call \procedurename~\ref{Procedure:NewtonRaphson1} to update $\boldsymbol{\alpha}^{(A)}_{1:K}$\;
  Call \procedurename~\ref{Procedure:NewtonRaphson1} to update $\boldsymbol{\beta}_{1:K}$\;

\caption{Mstep($q, \boldsymbol{\alpha}^{(\pi)}, \boldsymbol{\alpha}^{(A)}_{1:K}, \boldsymbol{\beta}_{1:K}$)}\label{Procedure:Mstep}
\end{procedure}

\subsubsection{The PF Form}
The process is the same as the above process.



\section{Comparison to Related Models} \label{Section:RelatedModels}
In this section, we compare our proposed LDHMMs to existing models that can model sequential behaviors, and our aim is to show the key difference between them.

\subsection{Hidden Markov Models}
As mentioned in \sectionname~\ref{Section:Introduction}, HMMs drop the difference of parameters between the sequences and assume all the sequences share the same parameters $\boldsymbol{\pi}$, $\mathbf{A}$ and $\mathbf{B}$, as shown in their graphical representation in \figurename~\ref{Figure:HMMs}. By contrast, LDHMMs assume the parameters representing the dynamics for each sequence are different and add hyper-parameters over the above parameters to avoid over-fittings.


\subsection{Latent Dirichlet Allocation}
Latent Dirichlet Allocation (LDA) \cite{Blei2003Latent} is a generative probabilistic model for a set of discrete data, which can be used for modeling sequential behaviors considered in this paper. Its graphical representation is shown in \figurename~\ref{Figure:LDA}. It can be seen from the figure that LDA simply ignores the dynamics in the hidden state space while LDHMMs consider the Markov relationships between hidden states.

\subsection{Variational Bayesian HMMs (VBHMMs)}
The graphical model of VBHMMs proposed in \cite{Beal2003Variational,Mckay1997Ensemble} is shown in \figurename~\ref{Figure:VBHMM}. Its difference to the proposed\linebreak LDHMMs is listed below: firstly, the VBHMMs assume all the sequences share the same parameters $\boldsymbol{\pi}$, $\mathbf{A}$ and $\mathbf{B}$ while LDHMMs assume each sequence has different parameters $\boldsymbol{\pi}_{m}$, $\mathbf{A}_{m}$ and $\mathbf{B}_{m}$ to characterize its dynamics; secondly, although VBHMMs use a variational inference algorithm similar to the PF form inference method described in \sectionname~\ref{Appendix:VI-PF} to learn the posterior distribution of the parameters, they assume the hyper-parameters are known and do not provide any algorithm for learning them. By contrast, the proposed LDHMMs adopt a variational EM framework to learn the hyper-parameters as well as the posterior distribution of the parameters.


\subsection{A Hidden Markov Model Variant (HMMV)}
\begin{figure}[!t]
\begin{center}
\subfigure[]{\label{Figure:HMMs}{\includegraphics[scale = 0.4]{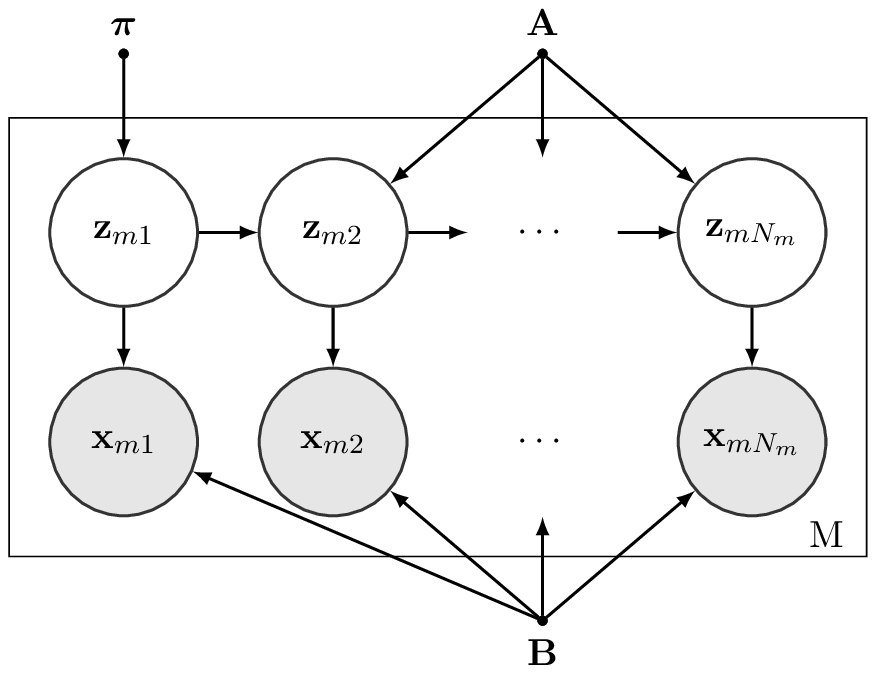}}}
\subfigure[]{\label{Figure:LDA}{\includegraphics[scale = 0.4]{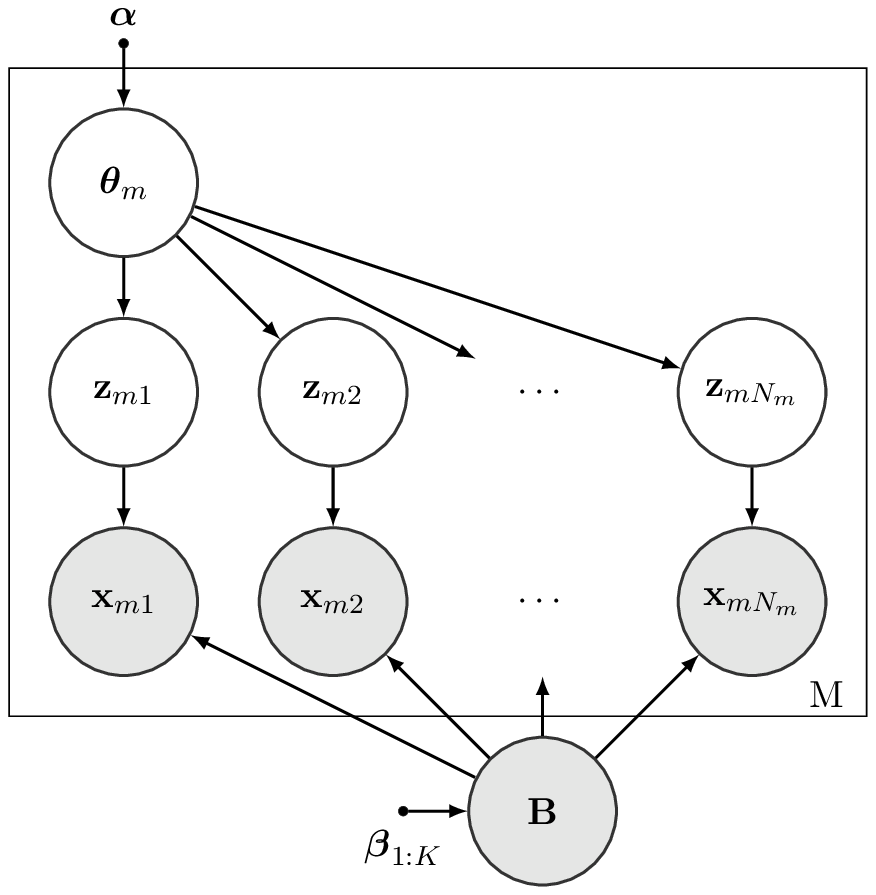}}}
\subfigure[]{\label{Figure:VBHMM}{\includegraphics[scale = 0.4]{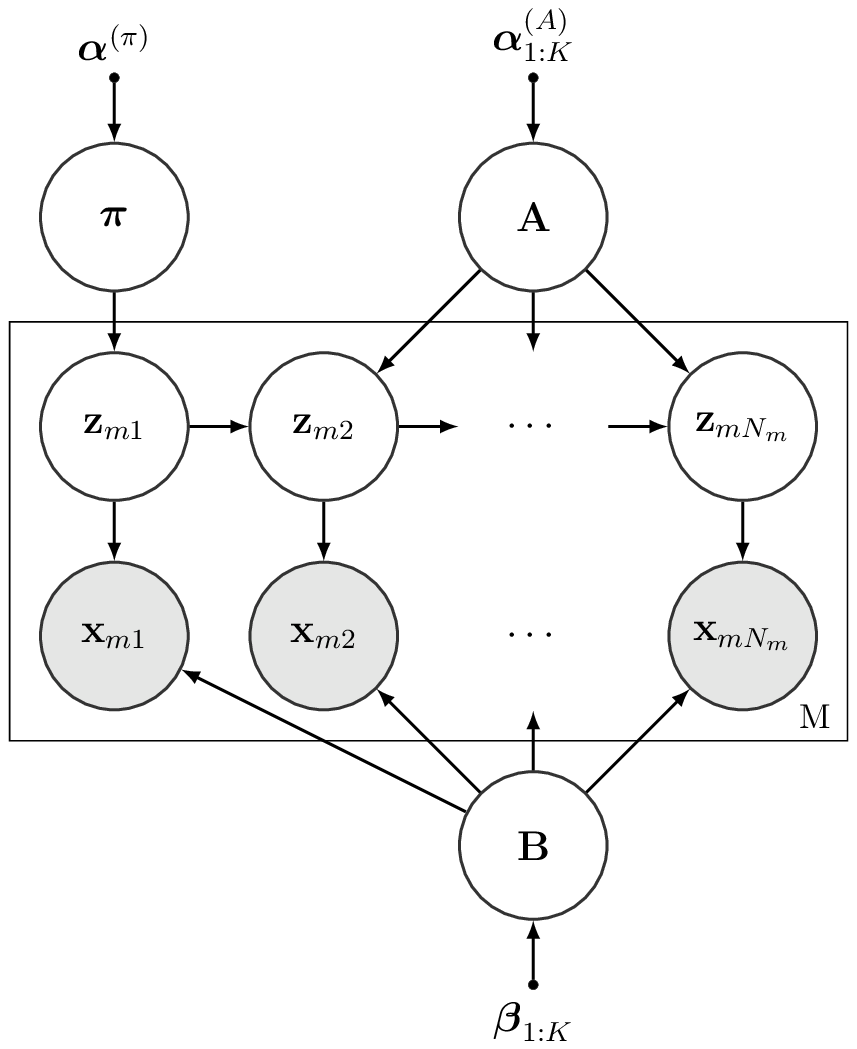}}}
\subfigure[]{\label{Figure:HMMV}{\includegraphics[scale = 0.4]{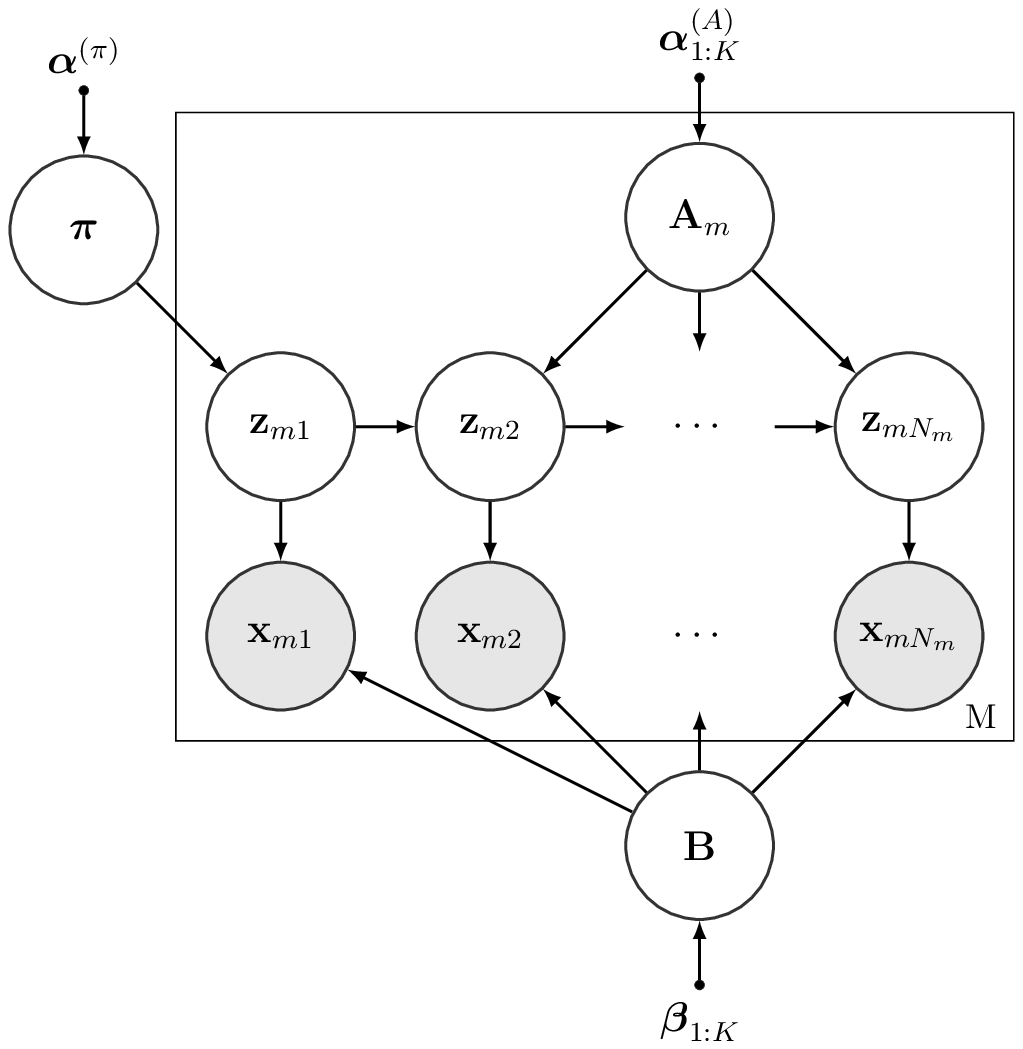}}}
\end{center}
\caption{The Graphical Models for (a) HMMs, (b) LDA, (c) VBHMMs (d) HMMV.}
\label{Figure:RelatedModels}
\end{figure}

As shown in \figurename~\ref{Figure:HMMV}, HMMV \cite{Blasiak2011A} has very similar graphical model compared to LDHMMs. However, it assumes  sequences share the same parameters $\boldsymbol{\pi}$ and $\mathbf{B}$. By contrast, LDHMMs treat these parameters individually for each sequence, in order to capture individual characteristics of them in a comprehensive way. Another difference to note is the assumption of known hyper-parameters. This is different from LDHMMs, since we provide a variational EM-based algorithm to estimate the hyper-parameters, as mentioned earlier.

\section{Empirical Study} \label{Section:EmpiricalStudy}
In this section, we apply the proposed LDHMMs in several data mining tasks, such as sequential behavior modeling and sequence classification. To be more specific, firstly, we use two public-available data sets from web-browsing logs to study the data mining tasks. Secondly, we adopt a public-available biological sequence data set to study the problem of sequence classification. All algorithms were implemented in matlab\footnote{The code will be made publicly available on the first author's web site soon.} and performed on a 2.9GHz 20MB L3 Cache Intel Xeon E5-2690 (8 Cores) cluster node with 32GB 1600MHz ECC DDR3-RAM (Quad Channel), running on a Red Hat Enterprise Linux 6.2 (64bit) operating system.

\subsection{Sequential Behavior Modeling}
\subsubsection{Data Sets}
\paragraph{The Entree Data Set}This data set\footnote{Available at http://archive.ics.uci.edu/ml/datasets/\linebreak Entree+Chicago+Recommendation+Data} records users' interactions with the Entree Chicago restaurant recommendation system from September, 1996 to April, 1999. The sequential behaviors of each user are his/her interactions with the system, i.e. their navigation operations. The characters L-T encode 8 navigation operations as shown in \tablename~\ref{Table:Entree}. We use a subset of 422 sequences whose lengths vary from 20 to 59.

\begin{table}[!t]
\renewcommand{\arraystretch}{1}
\caption{The Codebook of Navigation Operations}
\label{Table:Entree}
\begin{center}
\begin{tabular}{|c|c|}
  \hline
  Code & Navigation Operations \\ \hline \hline
  L & browse from one restaurant to another in a list \\ \hline
  M & search for a similar but cheaper restaurant\\ \hline
  N & search for a similar but nicer one\\ \hline
  P & search for a similar but more traditional one\\ \hline
  Q & search for a similar but more creative one\\ \hline
  R & search for a similar but more lively one\\ \hline
  S & search for a similar but quieter one\\ \hline
  T & search for a similar but different cuisine one\\ \hline
\end{tabular}
\end{center}
\end{table}

\paragraph{The MSNBC Data Set}This data set\footnote{Available at http://archive.ics.uci.edu/ml/datasets/\linebreak MSNBC.com+Anonymous+Web+Data.} describes the page visits of users who visited msnbc.com on September 28, 1999. Visits are recorded at the level of URL category (The 17 categories are `frontpage', `news', `tech', `local', `opinion', `on-air', `misc', `weather', `health', `living', `business', `sports', `summary', `bbs' (bulletin board service), `travel', `msn-news', and `msn-sports'.) and are recorded in a temporal order. Each sequence in the data set corresponds to page viewing behaviors of a user during that twenty-four hour period. Each behavior recorded in the sequence corresponds to the category of the user's requesting page. We use a subset of 31071 sequences whose lengths vary from 20 to 100.

\subsubsection{Evaluation Metrics}
We learned the LDHMMs of the proposed two different learning forms (denoted as LDHMMs-ff and LDHMMs-pf) and related models (i.e., HMMs, LDA, VBHMMs and HMMV), on the above two data sets to compare the generalization performance of these models. Our goal is to achieve high likelihood on a held-out test set. Thus, we computed the log-likelihood of a held-out test set to evaluate the models given the learned deterministic hyper-parameters/parameters. In particular, for LDHMMs, we first learned their deterministic database-level hyper-parameters according to \algorithmname~\ref{Algorithm:LDHMMs} using the training data; then approximately inferred the sequence-level parameters of the testing data by applying \procedurename~\ref{Procedure:Estep-ff} with the learned hyper-parameters; and finally computed the log-likelihood of the test data as \equationname~\ref{Equation:LowerBound2Expanded} using the learned hyper-parameters and inferred parameters. For other models, we used similar processes adjusting to their learning and inference algorithms. A higher log-likelihood indicates better generalization performance. We performed 10-fold cross validation on the above two data sets. Specifically, we split the data into 10 folds. Each time we held out 1 fold of the data for testing and trained the models on the remained 9-folds, and this process was repeated for 10 times. We report the averaged results of the 10-fold cross validation in the following.

\subsubsection{Comparison of Log-likelihood on the Test Data Set}
\begin{figure}[!t]
\begin{center}
\includegraphics[scale = 0.5]{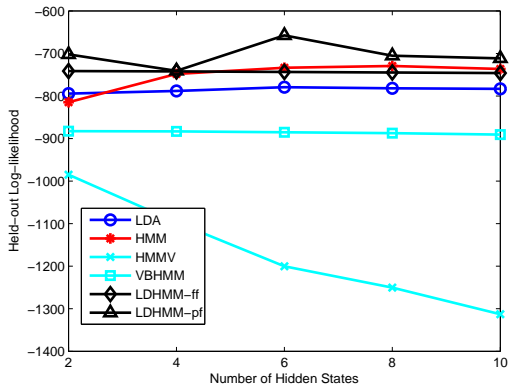}
\end{center}
\caption{Log-likelihood Results on the Entree Data Set for the Models.}
\label{Figure:Entree}
\end{figure}

\begin{figure}[!t]
\begin{center}
\includegraphics[scale = 0.5]{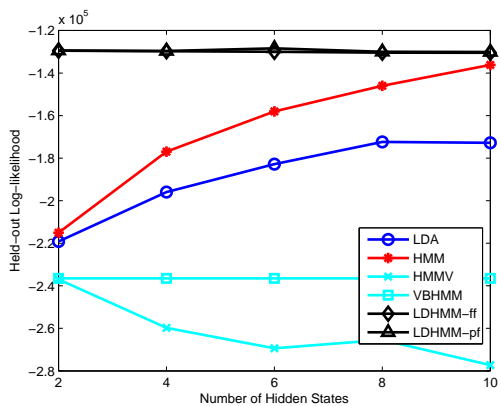}
\end{center}
\caption{Log-likelihood Results on the MSNBC Data Set for the Models.}
\label{Figure:MSNBC}
\end{figure}

 The results for different number of hidden states $K$ on the Entree data set is shown in \figurename~\ref{Figure:Entree}. As seen from the chart, the LDHMMs-pf consistently performs the best and LDHMMs-ff generally has the second best performance (only slightly worse than HMMs sometimes). Similar trend can be observed in \figurename~\ref{Figure:MSNBC}. Both LDHMMs-ff and LDHMMs-pf perform better than the other models while LDHMMs-pf has a slightly better performance. This is because the PF form may have a more accurate approximation in these data sets. In summary, the proposed LDHMMs has a better generalization performance compared to other models. To further validate the statistical significance of our experiments, we also perform the paired t-test (2-tail) between LDHMMs-pf, LDHMMs-ff and other models over the perplexity of the experimental results. All the t-test results are less than 0.01, which proves the improvements of LDHMMs over other models are statistically significant.


\subsubsection{Comparison of Computational time for the two forms}
\begin{figure}[!t]
\begin{center}
\subfigure[]{\label{Figure:Entree-t}{\includegraphics[scale = 0.25]{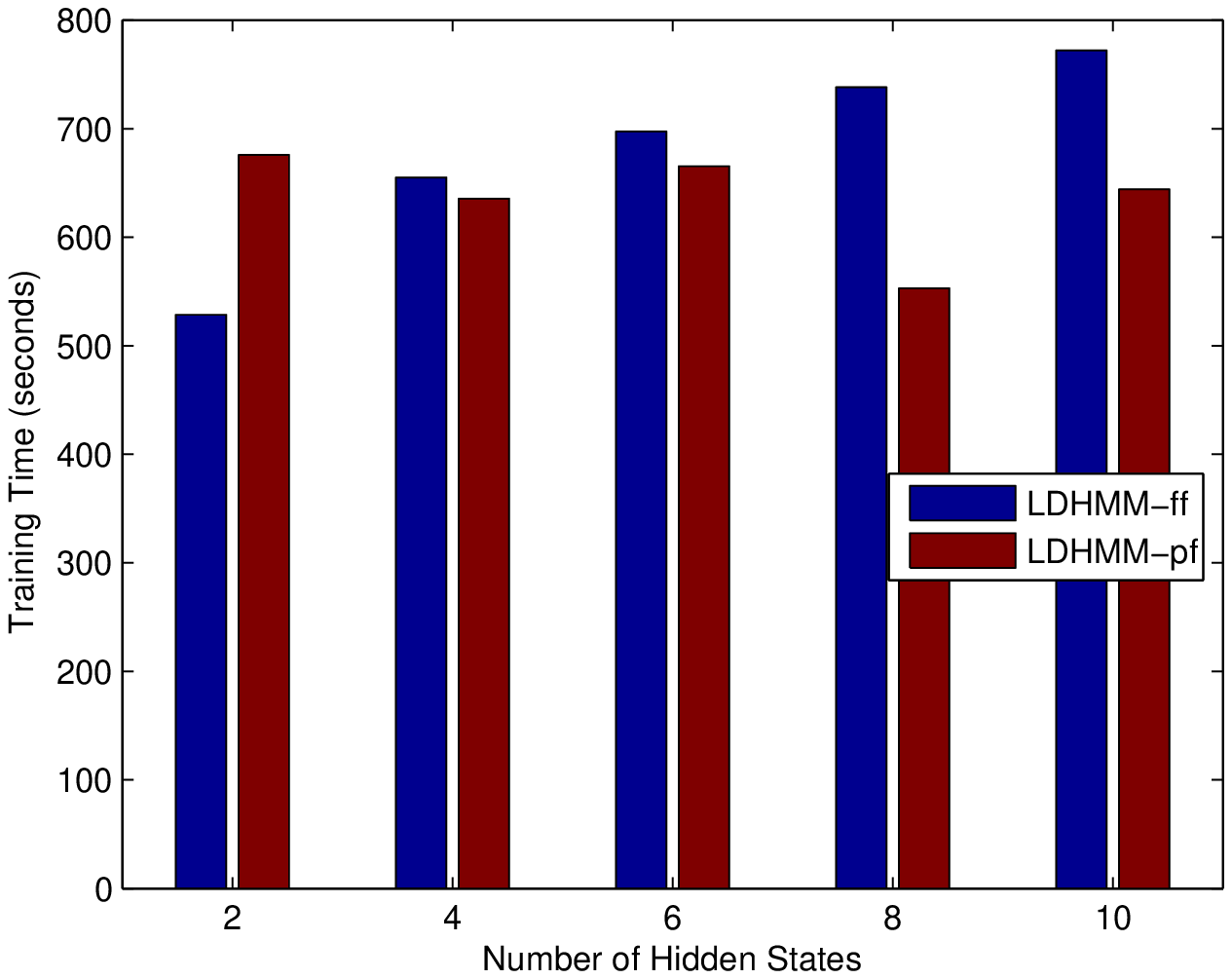}}} \hspace{-0.1em}
\subfigure[]{\label{Figure:MSNBC-t}{\includegraphics[scale = 0.25]{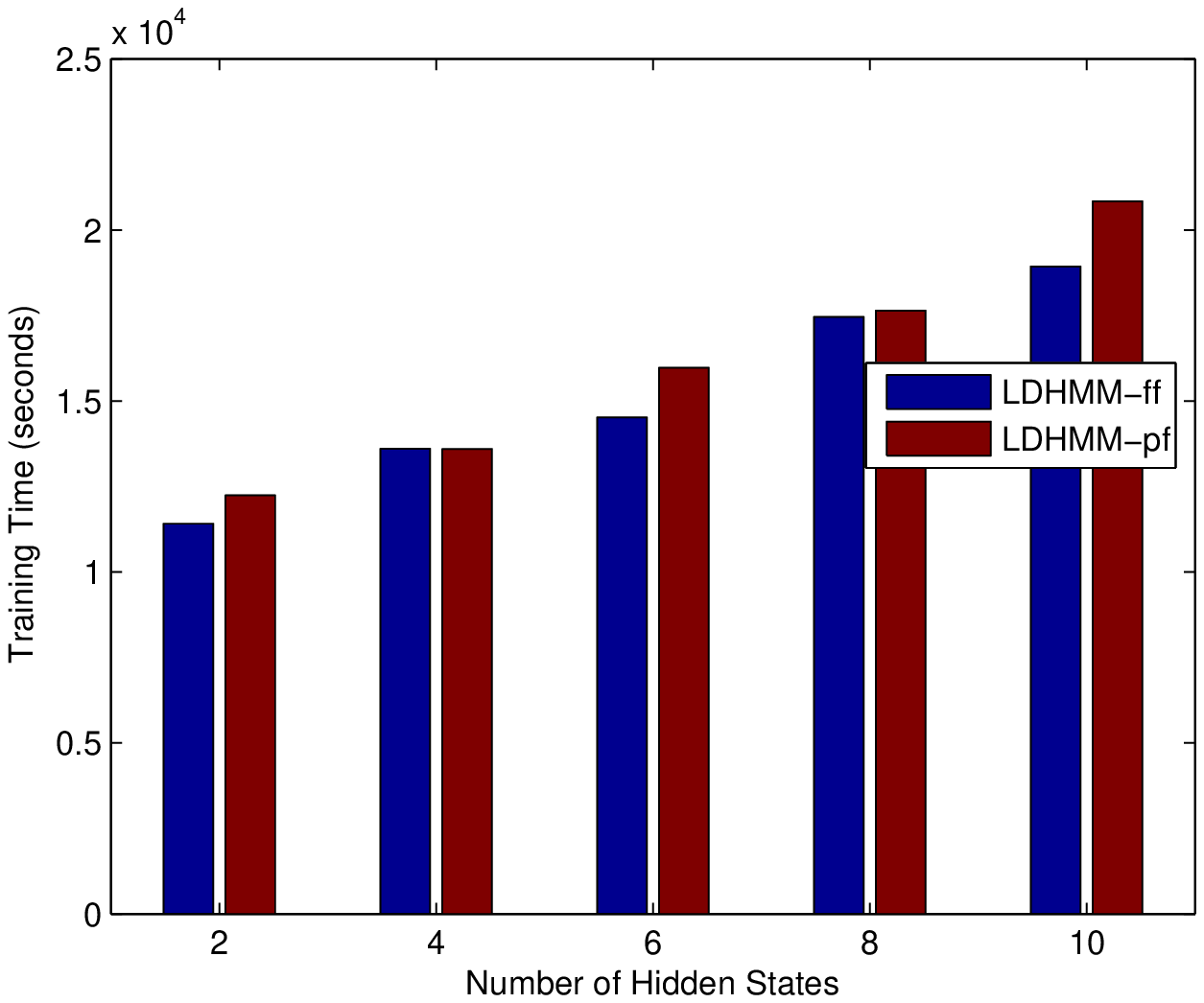}}}
\end{center}
\caption{Comparison of Training Time for the LDHMMs on (a) Entree Data Set, (b) MSNBC Data Set.}
\label{Figure:CompareTime}
\end{figure}

Since the computational complexity of related models are much lower than the proposed model due to their simpler structures, here we focus on the comparison of the proposed two forms. \figurename~\ref{Figure:CompareTime} shows the comparison of training time on the two used data sets. Qualitatively speaking, the two approaches have similar computational time. But sometimes, the PF form is faster the FF form, which seems to be contradict to our theoretical analysis in \sectionname~\ref{Subsubsection:Complexity}. However, in practice, the stopping criterion used in the EM algorithm which may cause the iteration to stop earlier. Since the PF form may converge faster than the FF form does, it may need less numbers of E and M steps. Thus, it may converge faster than the FF form in those cases.

%

\subsubsection{Visualization of LDHMMs}
\begin{figure}[!t]
\begin{center}
\subfigure{\label{Figure:VisualizeDatabase1}{\includegraphics[scale = 0.25]{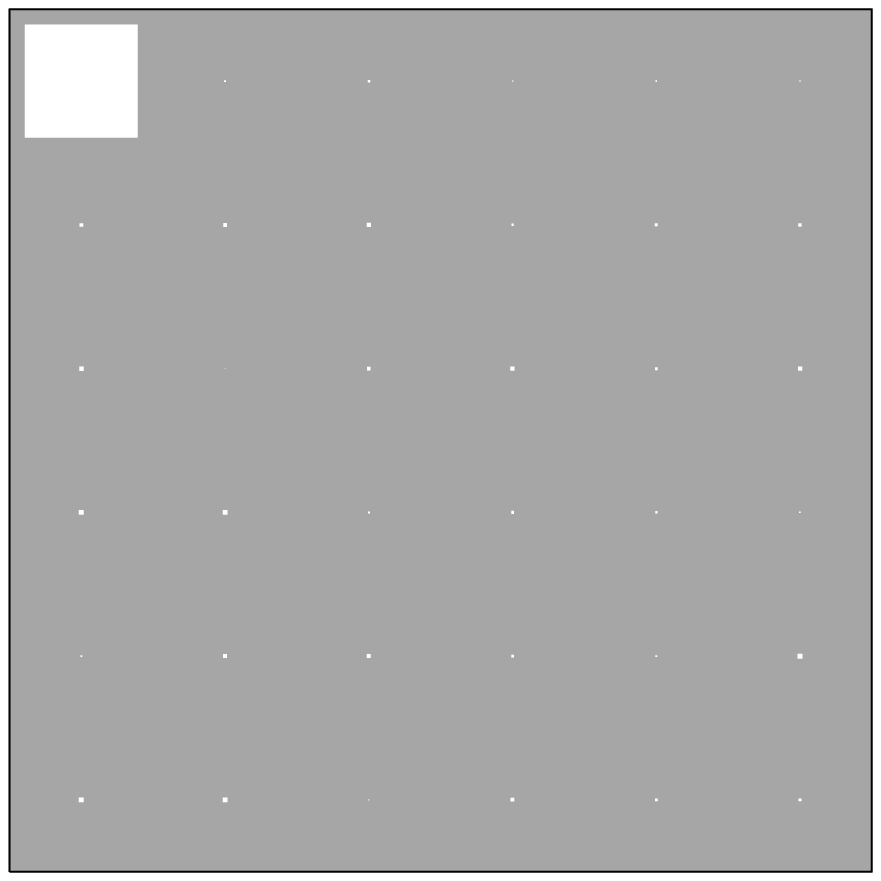}}} 
\subfigure{\label{Figure:VisualizeSeq1}{\includegraphics[scale = 0.25]{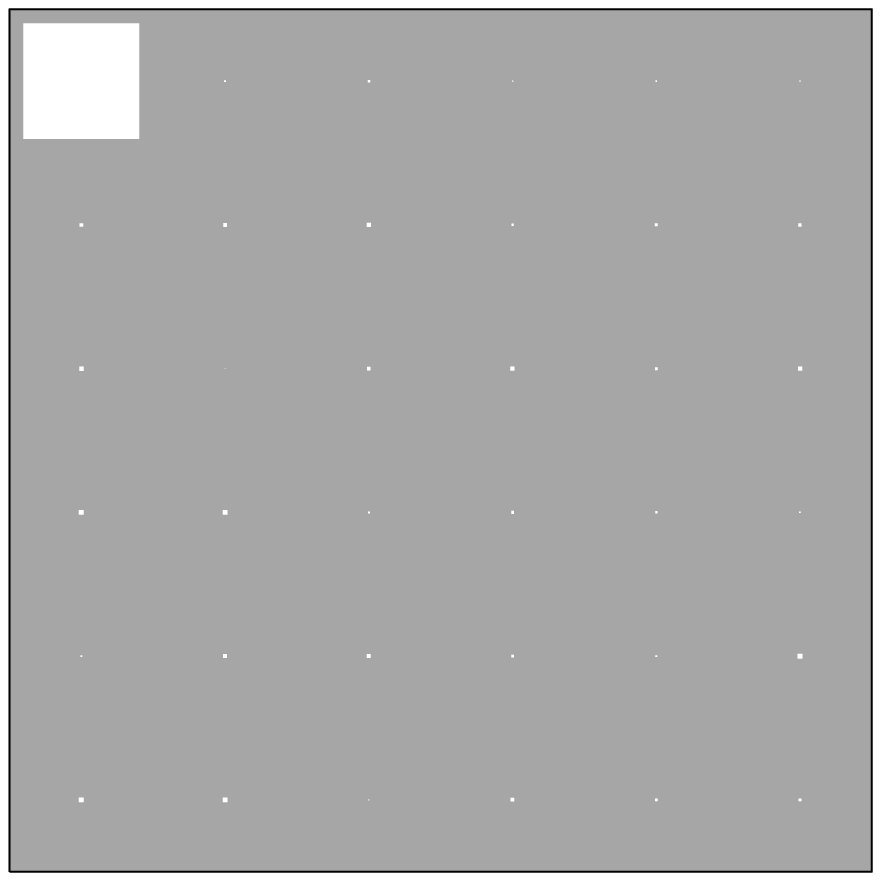}}}
\subfigure{\label{Figure:VisualizeDatabase2}{\includegraphics[scale = 0.25]{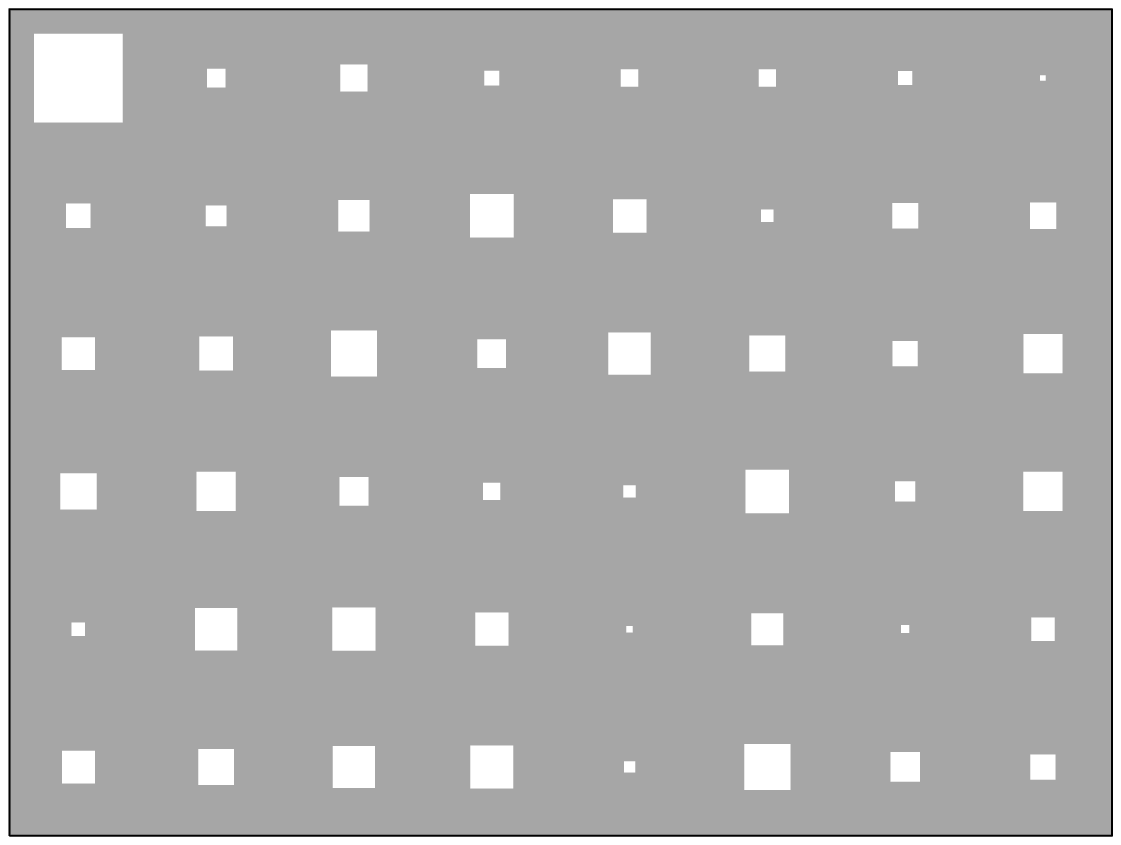}}} 
\subfigure{\label{Figure:VisualizeSeq2}{\includegraphics[scale = 0.25]{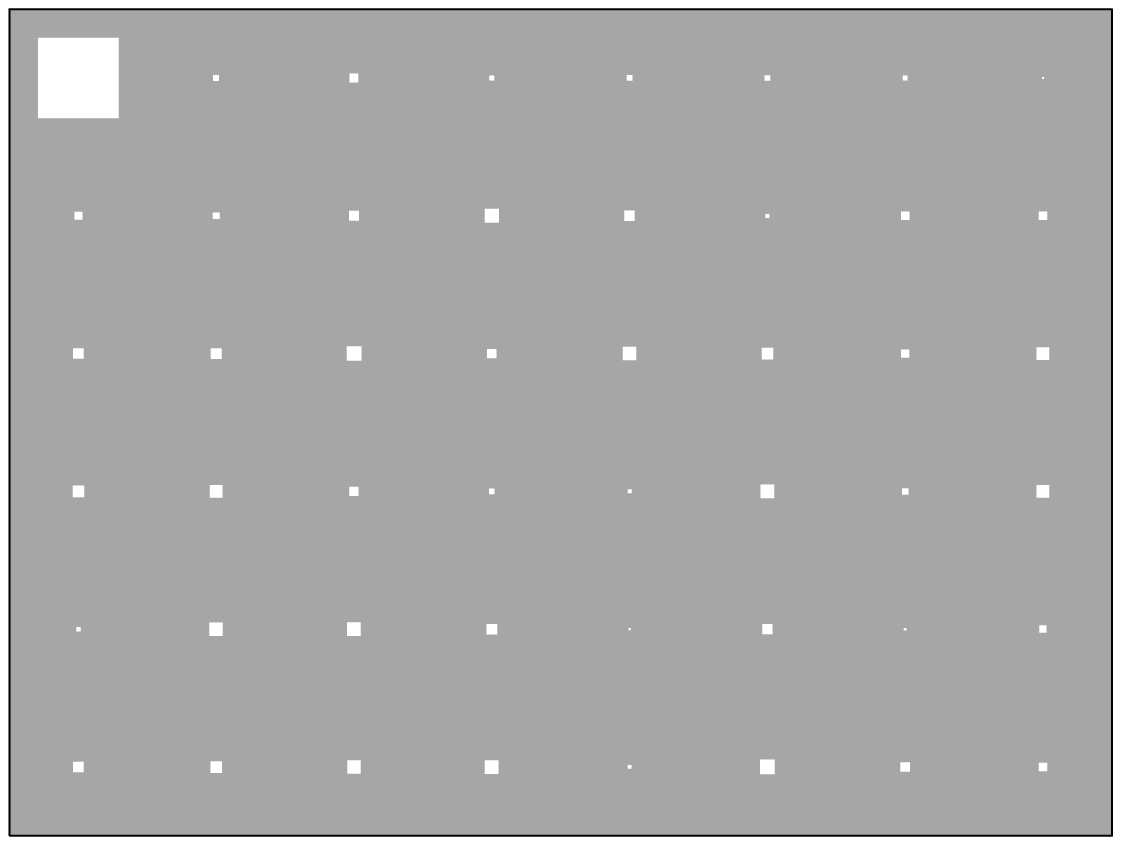}}}
\subfigure{\label{Figure:VisualizeDatabase3}{\includegraphics[scale = 0.25]{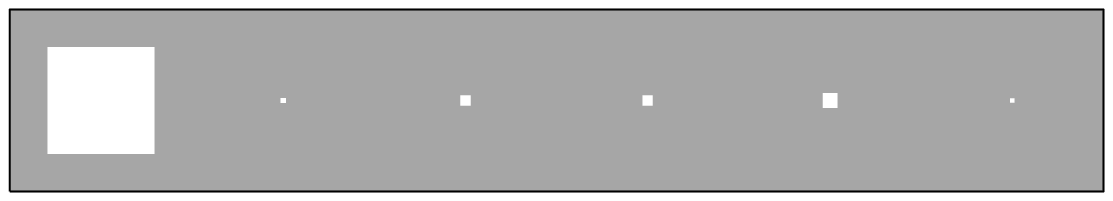}}} 
\subfigure{\label{Figure:VisualizeSeq3}{\includegraphics[scale = 0.25]{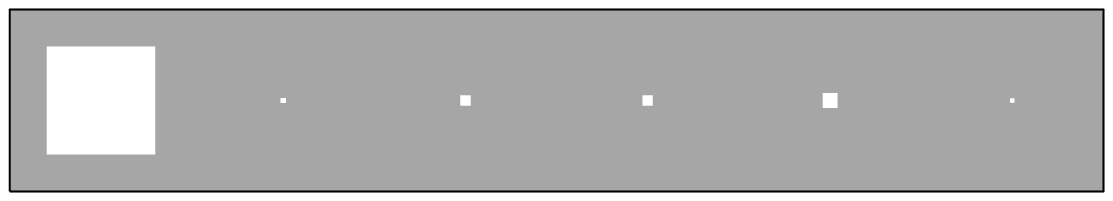}}}
\end{center}
\caption{The Hinton Diagrams for (a) Database Level Parameters, (b) Sequence Level Variational Parameters.}
\label{Figure:Visulization}
\end{figure}

It is also important to obtain an intuitive understanding of the complex model learned. LDHMMs have database-level hyper-parameters, (i.e., $\boldsymbol{\alpha}^{(\pi)}$, $\boldsymbol{\alpha}^{(A)}_{1:K}$ and $\boldsymbol{\beta}_{1:K}$), which can be seen as database-level characteristics of the sequences; sequence-level variational parameters (i.e., $\boldsymbol{\gamma}^{(\pi)}_{m}$, $\boldsymbol{\gamma}^{(B)}_{m, 1:K}$ and $\boldsymbol{\gamma}^{(A)}_{m, 1:K}$), which can be seen as sequence-level characteristics of each individual sequence. To visualize LDHMMs, we plot Hinton Diagrams for these parameters, each of which is represented by a square whose size is associated with the magnitude. \figurename~\ref{Figure:Visulization} shows a sample visualization from the Entree data set when $K = 6$. The left diagrams represent $\boldsymbol{\alpha}^{(A)}_{1:K}$, $\boldsymbol{\beta}_{1:K}$ and $\boldsymbol{\alpha}^{(\pi)}$ from the top to the bottom; the right diagrams represent $\boldsymbol{\gamma}^{(B)}_{m, 1:K}$, $\boldsymbol{\gamma}^{(A)}_{m, 1:K}$ and $\boldsymbol{\gamma}^{(\pi)}_{m}$ from the top to the bottom for a sample sequence from the data set. It is clear from the picture that the individual sequence displays slightly different characteristics from the whole database.

\subsection{Sequence Classification}
\subsubsection{Data Sets}
This data set\footnote{Available at http://archive.ics.uci.edu/ml/datasets/Molecular\linebreak \mbox{+Biology+\%28Splice-junction+Gene+Sequences\%29}} consists of 3 classes of DNA sequences. One class is made up of 767 sequences belonging to the exon/intron boundaries, referred to as the EI class; another class of 768 sequences belongs to the intron/exon boundaries, referred to as the IE class; the third class of 1655 sequences does not belong to either of the above classes, referred to as the N class.

\subsubsection{Evaluation Metrics}
We conducted 3 binary classification experiments (i.e., EI vs IE, EI vs N and IE vs N) using 10-fold cross validation. For each class $c$, we learned a separate model $p(\mathbf{X}_{1:M_{c}} | c)$ of the sequences in that class, where $M_{c}$ is the number of training sequences of class $c$. An unseen sequence was classified by picking $\arg\max_{c} p(\mathbf{X} | c) p(c)$. To eliminate the influence of $p(c)$, we varied its value and obtained the corresponding area under ROC curve (AUC) \cite{Fawcett2006An}, which is widely used for classification performance comparison.

\subsubsection{Comparison of AUC on the Test Data Set}
\tablename~\ref{Table:ClassificationResult} reports the averaged results on the 10-fold validation and the best results for each number of hidden states are in bold. Surprisingly, our proposed LDHMMs do not significantly dominate other models. An possible explanation is that the generative models are not optimized for classification and thus more accurate modeling does not result in the significant improvement of classification performance. This problem may be alleviated by combining the model with a discriminative classifier, which will be further discussed in \sectionname~\ref{Section:Conclusions}. However, LDHMMs have very competitive performance compared to the best models in all cases.

\begin{table*}
\renewcommand{\arraystretch}{1.3}
\caption{The Experimental Results of the Real Data Sets}
\label{Table:ClassificationResult}
\begin{center}
\begin{tabular}{c | c | c c c c c c}
\hline
Dataset& $Q$ &\textbf{LDA}&\textbf{HMM}&\textbf{HMMV}&\textbf{VBHMM}&\textbf{LDHMM-ff}&\textbf{LDHMM-nf}\\\hline
\multirow{3}{*}{EI vs IE} & {2}&$0.829\pm0.032$&$\bf 0.838\pm0.037$&$0.829\pm0.034$&$0.829\pm0.034$&$0.828\pm0.035$&$0.834\pm0.031$\\
&{3}&$\bf 0.829\pm0.032$&$0.827\pm0.046$&$\bf 0.829\pm0.034$&$\bf 0.829\pm0.034$&$0.828\pm0.035$&$0.791\pm0.044$\\
&{4}&$\bf 0.829\pm0.032$&$0.81\pm0.048$&$\bf 0.829\pm0.034$&$\bf 0.829\pm0.034$&$0.828\pm0.035$&$0.803\pm0.048$\\\hline
\multirow{3}{*}{EI vs N} & {2}&$0.67\pm0.029$&$0.679\pm0.038$&$0.67\pm0.029$&$0.67\pm0.029$&$0.677\pm0.03$&$\bf 0.689\pm0.03$\\
&{3}&$0.667\pm0.026$&$0.649\pm0.033$&$0.669\pm0.029$&$0.67\pm0.029$&$\bf 0.677\pm0.03$&$0.671\pm0.021$\\
&{4}&$0.671\pm0.025$&$0.659\pm0.036$&$0.67\pm0.029$&$0.67\pm0.029$&$0.677\pm0.03$&$\bf 0.678\pm0.03$\\ \hline
\multirow{3}{*}{IE vs N} &{2}&$0.724\pm0.036$&$0.739\pm0.029$&$0.724\pm0.036$&$0.724\pm0.036$&$0.734\pm0.033$&$\bf 0.743\pm0.028$\\
&{3}&$0.725\pm0.034$&$0.66\pm0.024$&$0.724\pm0.036$&$0.724\pm0.036$&$\bf 0.734\pm0.033$&$0.733\pm0.032$\\
&{4}&$0.729\pm0.033$&$0.721\pm0.023$&$0.723\pm0.036$&$0.724\pm0.036$&$\bf 0.734\pm0.033$&$0.725\pm0.039$\\\hline
\end{tabular}
\end{center}
\end{table*}

\section{Conclusions and Future Work} \label{Section:Conclusions}
Statistical modeling of sequential data has been studied for many years in machine learning and data mining. In this paper, we propose LDHMMs to characterize a database of sequential behaviors. Rather than assuming all the sequences share the same parameters as in traditional models, such as HMMs and VBHMMs, we explicitly assign sequence-level parameters to each sequence and database-level hyper-parameters to the whole database. The experimental results show that our model outperforms the other state-of-the-art models in predicting unseen sequential behaviors from web browsing logs and is competitive in classifying the unseen biological sequences.

In this paper, we assume that the observed sequences can only have one behavior at one time stamp, which is not practical in many application domains. For example, in the field of customer transaction analysis, one customer may buy several items at one time stamp. Thus, one possible future direction is to generalize LDHMMs to cater for the above scenarios. Additionally, it is also interesting to investigate the combination of our model with discriminative classifiers, such as support vector machine (SVM), to further improve the classification performance. This is because, similar to LDA, our model can be naturally seen as a dimensionality reduction method for feature extraction.

\section{Acknowledgements}
Thanks for the anonymous reviewers!

%
\bibliographystyle{abbrv}
\bibliography{./Bib/VB,./Bib/MLF,./Bib/HMM,./Bib/FPM,./Bib/TM}  
%
%
\appendix
\section{Distributions} \label{Appendix:Distributions}
\subsection{Dirichlet Distribution}
A $K$-dimensional Dirichlet random variable $\boldsymbol{\theta}$ ($\sum_{i = 1}^{K} \theta_{i} = 1$ and $\theta_{i} \ge 0$) and has the following form of probability distribution \cite{Kotz2000Continuous}:
\begin{equation}
  p(\boldsymbol{\theta}; \boldsymbol{\alpha}) = \frac{\Gamma (\sum_{i = 1}^{K}) \alpha_{i}}{\prod_{i = 1}^{K} \Gamma (\alpha_{i})} \theta_{1}^{\alpha_{1} - 1} \cdots \theta_{K}^{\alpha_{K} - 1}
\end{equation}
where the parameter $\boldsymbol{\alpha}$ is a $K$-dimension vector with components $\alpha_{i} > 0$, and where $\Gamma (\cdot)$ is the Gamma function.
\subsection{Multinomial Distribution}
A 1-of-$V$ vector multinomial random variable $\boldsymbol{x}$ ($\sum_{i = 1}^{V} x_{i} = 1$ and $x_{i} \in \{0, 1\}$) and has the following form of probability distribution \cite{Evans2000Statistical}:
\begin{equation}
  p(\mathbf{x}; \boldsymbol{\mu}) = \prod_{i = 1}^{V} \mu_{i}^{x_{i}}
\end{equation}
where the parameter $\boldsymbol{\mu}$ is a $V$-dimension vector with components $\sum_{i = 1}^{K} \mu_{i} = 1$ and $\mu_{i} \ge 0$.

\section{Variational Inference}
\subsection{The FF Form} \label{Appendix:VI-FF}
Here we expand the expression of $L$ for Variational Inference for the FF form. for $q_{m}(\boldsymbol{\pi}_{m}; \boldsymbol{\gamma}^{(\pi)}_{m})$, $q_{m}(\mathbf{A}_{m}; \boldsymbol{\gamma}^{(A)}_{m, 1:K})$ and $q_{m}(\mathbf{B}_{m}; \boldsymbol{\gamma}^{(B)}_{m, 1:K})$ are usually assumed to be Dirichlet distributions governed by parameters $\boldsymbol{\gamma}^{(\pi)}_{m}$, $\boldsymbol{\gamma}^{(A)}_{m, 1:K}$ and $\boldsymbol{\gamma}^{(B)}_{m, 1:K}$. $q_{m}(\mathbf{z}_{m n}; \boldsymbol{\phi}_{m n})$ is assumed to be multinomial distributions governed by $\boldsymbol{\phi}_{m n}$ ($1 \le n \le N_{m}$). Then \equationname~\ref{Equation:LowerBound} can be expanded as follows:
\begin{equation}  \begin{array}{l}\label{Equation:LowerBound2Expanded}
     L = \sum_{m = 1}^{M} [
           E_{q} [ \log p(\boldsymbol{\pi}_{m} | \boldsymbol{\alpha}^{(\pi)}) ]
           + E_{q} [ \log p(\mathbf{A}_{m} | \boldsymbol{\alpha}^{(A)}_{1:K}) ] \nonumber \\
       + E_{q} [ \log p(\mathbf{B}_{m} | \boldsymbol{\beta}_{1:K}) ]
           + E_{q} [ \log p(\mathbf{z}_{m1} | \boldsymbol{\pi}_{m}) ] \nonumber \\
       + E_{q} [ \sum_{n = 2}^{N_{m}} \log p(\mathbf{z}_{m n} | \mathbf{z}_{m, n-1}, \mathbf{A}_{m}) ] \nonumber \\
       + E_{q} [ \sum_{n = 1}^{N_{m}} \log p( \mathbf{x}_{m n}| \mathbf{z}_{m n}, \mathbf{B}_{m})] \nonumber \\
       - E_{q} [ \log q_{m}(\boldsymbol{\pi}_{m}) ] - E_{q} [ \log q_{m}(\mathbf{A}_{m}) ] \nonumber \\
       - E_{q} [ \log q_{m}(\mathbf{B}_{m}) ]
         - E_{q} [ \log q_{m}(\mathbf{Z}_{m}) ]
         ]
         \end{array}
\end{equation}

where
\begin{equation}
\begin{array}{l}
   E_{q} [ \log p(\boldsymbol{\pi}_{m} | \boldsymbol{\alpha}^{(\pi)}) ] = \log \Gamma(\sum_{j = 1}^{K} \alpha^{(\pi)}_{j}) - \sum_{i = 1}^{K} \log \Gamma(\alpha^{(\pi)}_{i}) \\
   + \sum_{i = 1}^{K} (\alpha^{(\pi)}_{i} - 1) (\Psi (\gamma^{(\pi)}_{m i}) - \Psi (\sum_{j = 1}^{K} \gamma^{(\pi)}_{m j}))
\end{array}
\end{equation}
\begin{equation} \begin{array}{l}
   E_{q} [ \log p(\mathbf{A}_{m} | \boldsymbol{\alpha}^{(A)}_{1:K}) ] = \sum_{i = 1}^{K} [ \log \Gamma(\sum_{j = 1}^{K} \alpha^{(A)}_{i j}) - \sum_{k = 1}^{K} \\ \log \Gamma(\alpha^{(A)}_{i k})
   + \sum_{k = 1}^{K} (\alpha^{(A)}_{i k} - 1) (\Psi (\gamma^{(A)}_{m i k}) - \Psi (\sum_{j = 1}^{K} \gamma^{(A)}_{m i j})) ]
\end{array} \end{equation}
\begin{equation}
  \begin{array}{l}
   E_{q} [ \log p(\mathbf{B}_{m} | \boldsymbol{\beta}_{1:K}) ] = \sum_{i = 1}^{K} [ \log \Gamma(\sum_{j = 1}^{K} \beta_{i j}) - \sum_{k = 1}^{K} \\ \log \Gamma(\beta_{i k})
   + \sum_{k = 1}^{K} (\beta_{i k} - 1) (\Psi (\gamma^{(B)}_{m i k}) - \Psi (\sum_{j = 1}^{K} \gamma^{(B)}_{m i j})) ]
  \end{array}
\end{equation}
\begin{equation}  \label{Equation:item4}
   E_{q} [ \log p(\mathbf{z}_{m1} | \boldsymbol{\pi}_{m}) ] = \sum_{i = 1}^{K} \phi_{m1i} (\Psi (\gamma^{(\pi)}_{m i}) - \Psi (\sum_{j = 1}^{K} \gamma^{(\pi)}_{m j}))
\end{equation}
\begin{equation} \begin{array}{l}  \label{Equation:item5}
   E_{q} [ \sum_{n = 2}^{N_{m}} \log p(\mathbf{z}_{m n} | \mathbf{z}_{m, n-1}, \mathbf{A}_{m}) ] = \sum_{n = 2}^{N} \sum_{i = 1}^{K} \sum_{k = 1}^{K} \\ \phi_{m n-1, i} \phi_{m n k}
   (\Psi (\gamma^{(A)}_{m i k}) - \Psi (\sum_{j = 1}^{K} \gamma^{(A)}_{m i j}))
\end{array} \end{equation}
\begin{equation} \begin{array}{l}  \label{Equation:item6}
   E_{q} [ \sum_{n = 1}^{N_{m}} \log p( \mathbf{x}_{m n}| \mathbf{z}_{m n}, \mathbf{B}_{m})] =  \sum_{n = 1}^{N_{m}} \sum_{i = 1}^{K} \sum_{j = 1}^{V}  \phi_{m n i} \\
   (x_{m n j} ( \Psi (\gamma^{(B)}_{m i j}) - \Psi (\sum_{v = 1}^{V} \gamma^{(B)}_{m i v})))
\end{array} \end{equation}
\begin{equation} \begin{array}{l}
   E_{q} [ \log q_{m}(\boldsymbol{\pi}_{m}) ] = \log \Gamma(\sum_{j = 1}^{K} \gamma^{(\pi)}_{m j}) - \sum_{i = 1}^{K} \log \Gamma(\gamma^{(\pi)}_{m i}) \\
   + \sum_{i = 1}^{K} (\gamma^{(\pi)}_{m i} - 1) (\Psi (\gamma^{(\pi)}_{m i}) - \Psi (\sum_{j = 1}^{K} \gamma^{(\pi)}_{m j}))
\end{array} \end{equation}
\begin{equation} \begin{array}{l}
   E_{q} [ \log q_{m}(\mathbf{A}_{m}) ] =  \sum_{i = 1}^{K} [ \log \Gamma(\sum_{j = 1}^{K} \gamma^{(A)}_{m i j}) - \sum_{k = 1}^{K} \\ \log \Gamma(\gamma^{(A)}_{m i k}) +
   \sum_{k = 1}^{K} (\gamma^{(A)}_{m i k} - 1) (\Psi (\gamma^{(A)}_{m i k}) - \Psi (\sum_{j = 1}^{K} \gamma^{(A)}_{m i j})) ]
\end{array} \end{equation}
\begin{equation} \begin{array}{l}
   E_{q} [ \log q_{m}(\mathbf{B}_{m}) ] = \sum_{i = 1}^{K} [ \log \Gamma(\sum_{j = 1}^{V} \gamma^{(B)}_{m i j}) - \sum_{v = 1}^{V} \\\log \Gamma(\gamma^{(B)}_{m i v}) +
   \sum_{j = 1}^{V} (\gamma^{(B)}_{m i j} - 1) (\Psi (\gamma^{(B)}_{m i j}) - \Psi (\sum_{v = 1}^{V} \gamma^{(B)}_{m i v})) ]
\end{array} \end{equation}
\begin{equation}
    E_{q} [ \log q_{m}(\mathbf{Z}_{m}) ] =  \sum_{n = 1}^{N_{m}} \sum_{i = 1}^{K} \phi_{m n i} \log \phi_{m n i}
\end{equation}

\paragraph{Fixed $\boldsymbol{\gamma}^{(\pi)}_{m}$ and $\boldsymbol{\gamma}^{(B)}_{m, 1:K}$ and $\boldsymbol{\gamma}^{(A)}_{m, 1:K}$, Update $\boldsymbol{\phi}_{m, 1:N_{m}}$} \label{Appendix:VI-FF-updatephi}
As a functional of ${\phi}_{m1i}$ and add Lagrange multipliers:
\begin{equation} \begin{array}{l} \label{Equation:LowerBound-z1-ff}
 L(\boldsymbol{\phi}_{m1})  =  \sum_{i = 1}^{K} \phi_{m1i} (\Psi (\gamma^{(\pi)}_{m i}) - \Psi (\sum_{j = 1}^{K} \gamma^{(\pi)}_{m j})) \\
                             + \sum_{i = 1}^{K} \sum_{k = 1}^{K} \phi_{m 1 i} \phi_{m 2 k} (\Psi (\gamma^{(A)}_{m i k}) - \Psi (\sum_{j = 1}^{K} \gamma^{(A)}_{m i j})) \\
                             + \sum_{i = 1}^{K} \sum_{j = 1}^{V}  \phi_{m 1 i} (x_{m 1 j} ( \Psi (\gamma^{(B)}_{m i j}) - \Psi (\sum_{v = 1}^{V} \gamma^{(B)}_{m i v})))  \\
                             - \sum_{i = 1}^{K} \phi_{m 1 i} \log \phi_{m 1 i} + \lambda (\sum_{i = 1}^{K} \phi_{m 1 i} - 1) + const
\end{array} \end{equation}

Setting the derivative to zero yields the maximizing value of the variational parameter $\phi_{m 1 i}$ as \equationname~\ref{Equation:Update-z1-ff}. Similarly, the updated equation for $\phi_{m n i}$ ($2 \le n \le N_{m}-1$) and $\phi_{m N_{m} i}$ can be obtained as \equationname~\ref{Equation:Update-z2toNminus1-ff} and \ref{Equation:Update-zN-ff}.

Use similar technique as above, we can fix $\boldsymbol{\phi}_{m, 1:N_{m}}$, $\boldsymbol{\gamma}^{(A)}_{m, 1:K}$, $\boldsymbol{\gamma}^{(B)}_{m, 1:K}$, update $\boldsymbol{\gamma}^{(\pi)}_{m}$ as \equationname~\ref{Equation:Update-pi-ff}; fix $\boldsymbol{\phi}_{m, 1:N_{m}}$, $\boldsymbol{\gamma}^{(\pi)}_{m}$ and $\boldsymbol{\gamma}^{(B)}_{m, 1:K}$, update $\boldsymbol{\gamma}^{(A)}_{m, 1:K}$ as \equationname~\ref{Equation:Update-A-ff}; fix $\boldsymbol{\phi}_{m, 1:N_{m}}$, $\boldsymbol{\gamma}^{(\pi)}_{m}$ and $\boldsymbol{\gamma}^{(A)}_{m, 1:K}$, estimate $\boldsymbol{\gamma}^{(B)}_{m, 1:K}$ as \equationname~\ref{Equation:Update-B-ff}.

\subsection{The PF From}  \label{Appendix:VI-PF}
The expansion of $L$ for Variational Inference for the PF form is similar to the FF form except for changing $\phi_{m n i}$ to $\gamma_{m n i}$ ($1 \le n \le N_{m}$) in \equationname~\ref{Equation:item4} and \ref{Equation:item6}, and $\phi_{m, n-1, i} \phi_{m n k}$ to $\xi_{m, n-1, i, n, k}$ ($2 \le n \le N_{m}$) in \equationname~\ref{Equation:item5}, where
\begin{equation}
\begin{aligned}
   \gamma_{m n k} = p({z}_{m n k}| \mathbf{X}_{m}, \boldsymbol{\gamma}^{(\pi)}_{m}, \boldsymbol{\gamma}^{(B)}_{m, 1:K}, \boldsymbol{\gamma}^{(A)}_{m, 1:K}) \nonumber \\
   \xi_{m, n-1, j, n, k} = p({z}_{m, n-1, j}, {z}_{m n k} | \mathbf{X}_{m}, \boldsymbol{\gamma}^{(\pi)}_{m}, \boldsymbol{\gamma}^{(B)}_{m, 1:K}, \boldsymbol{\gamma}^{(A)}_{m, 1:K})
\end{aligned}
\end{equation}
Then, the variational inference can be done as following:
\paragraph{Fixed $\boldsymbol{\gamma}^{(\pi)}_{m}$ and $\boldsymbol{\gamma}^{(B)}_{m, 1:K}$ and $\boldsymbol{\gamma}^{(A)}_{m, 1:K}$, Update $q_{m}(\mathbf{Z}_{m})$} \label{Appendix:VI-PF-ForwardBackward}
As a functional of $q(\mathbf{z})$ the lower bound of log-likelihood can be expressed as follows:
\begin{equation} \begin{array}{l} \label{Equation:LowerBound-Z-nf}
   L(q_{m}(\mathbf{Z}_{m})) = \sum_{i = 1}^{K} \gamma_{m 1 i} (\Psi (\gamma^{(\pi)}_{m i}) - \Psi (\sum_{j = 1}^{K} \gamma^{(\pi)}_{m j})) \\
     + \sum_{n = 2}^{N_{m}} \sum_{i = 1}^{K} \sum_{k = 1}^{K} \xi_{m, n-1, i, n, k}) \\ (\Psi (\gamma^{(A)}_{m i k}) - \Psi (\sum_{j = 1}^{K} \gamma^{(A)}_{m i j})) + \sum_{n = 1}^{N_{m}} \sum_{i = 1}^{K} \sum_{j = 1}^{V}  \gamma_{m n i} \\ (x_{m n j} ( \Psi (\gamma^{(B)}_{m i j}) - \Psi (\sum_{v = 1}^{V} \gamma^{(B)}_{m i v}))) + const
\end{array} \end{equation}

Now, defining
\begin{equation} \begin{aligned}
  \boldsymbol{\pi}_{m}^{*} &\equiv& \exp (\Psi (\gamma^{(\pi)}_{m i}) - \Psi (\sum_{j = 1}^{K} \gamma^{(\pi)}_{m j})) \nonumber \\
  \mathbf{A}_{m}^{*} &\equiv&  \exp(\Psi (\gamma^{(A)}_{m i k}) - \Psi (\sum_{j = 1}^{K} \gamma^{(A)}_{m i j}))  \\
  \mathbf{B}_{m}^{*} &\equiv&  \exp ( \Psi (\gamma^{(B)}_{m i j}) - \Psi (\sum_{v = 1}^{V} \gamma^{(B)}_{m i v})))
\end{aligned} \end{equation}

The above form of $L$ is similar to the log-likelihood object function of standard HMMs \cite{Mckay1997Ensemble} and the relevant posteriors $\boldsymbol{\gamma}$, $\boldsymbol{\xi}$ can be calculated by the forward-backward (FB) algorithm \cite{Rabiner1990A}, which will be briefly reviewed in the following.

Here define the following auxiliary variables $\boldsymbol{\alpha}^{'}$ and $\boldsymbol{\beta}^{'}$ ($1 \le m \le M, 1 \le n \le N_{m}, 2 \le n^{'} \le N_{m}, 1 \le j \le K, 1 \le k \le K$ and $\boldsymbol{\theta}_{m}^{*} = \{ \boldsymbol{\pi}_{m}^{*}, \mathbf{A}_{m}^{*}, \mathbf{B}_{m}^{*} \}$):
\begin{equation}
\begin{aligned}
   \alpha_{m n k}^{'} = p(\mathbf{x}_{m 1}, \cdots, \mathbf{x}_{m n}, {z}_{m n k} | \boldsymbol{\theta}_{m}^{*}) \nonumber \\
   \beta_{m n k}^{'} = p(\mathbf{x}_{m, n+1}, \cdots, \mathbf{x}_{m N} | {z}_{m n k}, \boldsymbol{\theta}_{m}^{*})
\end{aligned}
\end{equation}
Then FB algorithm can be summarized in \algorithmname~\ref{Algorithm:ForwardBackward}. Specifically, line 1-5 calculate the forward variables $\boldsymbol{\alpha}^{'}$, while line 6-10 calculate the backward variables $\boldsymbol{\beta}^{'}$. Then line11-15 calculate the value of each element of the posteriors $\boldsymbol{\gamma}$ and $\boldsymbol{\xi}$ on the basis of the $\boldsymbol{\alpha}^{'}$, $\boldsymbol{\beta}^{'}$ and $\boldsymbol{\theta}^{*}_{m}$.

\begin{algorithm}[!t]
\SetKwInOut{In}{input}\SetKwInOut{Out}{output}
\In{An initial setting for the parameters $\boldsymbol{\theta}_{m}^{*}$}
\Out{Inferred posterior distributions $\boldsymbol{\gamma}$, $\boldsymbol{\xi}$}
\BlankLine
\tcc{Calculation of $\boldsymbol{\alpha}^{'}$, $\boldsymbol{\beta}^{'}$}
\tcp*[h]{Forward\;}
   $\alpha_{m 1 k}^{'} = \pi_{m k}^{*} p(\mathbf{x}_{m 1}; \mathbf{b}_{k})$ for $k$\;

\For(\tcp*[h]{Induction}){n=1 \emph{\KwTo}N-1}{
        \For{k=1 \emph{\KwTo}K}{
          $\alpha_{m, n+1, k}^{'} = \sum_{j=1}^{K} \alpha_{m n j}^{'} a_{j k}^{*} p(\mathbf{x}_{m, n+1}; \mathbf{b}_{k})$\;
        }
}
\tcp*[h]{Backward\;}
$\beta_{m N k}^{'} = 1$ for all $k$\;

\For(\tcp*[h]{Induction}){n=N-1 \emph{\KwTo}1}{
        \For{j=1 \emph{\KwTo}K}{
          $\beta_{m n k}^{'} = \sum_{j=1}^{K} a_{j k}^{*} p(\mathbf{x}_{m, n+1}; \mathbf{b}_{k}) \beta_{m, n+1, j}^{'}$\;
        }
}

\tcc{Calculation of $\boldsymbol{\gamma}$, $\boldsymbol{\xi}$}
      $p(\mathbf{X}_{m} | \boldsymbol{\theta}_{m}^{*}) = \sum_{k=1}^{K} \alpha_{s g m N k}$\;
      \For{n=1 \emph{\KwTo}N}{
        $\gamma_{m n k} = \frac{\alpha_{m n k}^{'} \beta_{m n k}^{'} }{p(\mathbf{X}_{m}| \boldsymbol{\theta}_{m}^{*})}$\;
        $\xi_{m, n-1, j, n, k} = \frac{\alpha_{m, n-1, k}^{'} p(\mathbf{x}_{m n}; \mathbf{b}_{k}) a_{j k}^{*} \beta_{m n k}^{'}}{p(\mathbf{X}_{m} | \boldsymbol{\theta}_{m}^{*})}$ ($n > 2$)\;

       }
\caption{ForwardBackward()}\label{Algorithm:ForwardBackward}
\end{algorithm}

Use similar techniques described in \appendixname~\ref{Appendix:VI-FF-updatephi}, we can fix $q_{m}(\mathbf{Z}_{m})$, $\boldsymbol{\gamma}^{(A)}_{m, 1:K}$, $\boldsymbol{\gamma}^{(B)}_{m, 1:K}$, update $\boldsymbol{\gamma}^{(\pi)}_{m}$ as \equationname~\ref{Equation:Update-pi-pf}; fix $q_{m}(\mathbf{Z}_{m})$, $\boldsymbol{\gamma}^{(\pi)}_{m}$ and $\boldsymbol{\gamma}^{(B)}_{m, 1:K}$, update $\boldsymbol{\gamma}^{(A)}_{m, 1:K}$ as \equationname~\ref{Equation:Update-A-pf}; fix $q_{m}(\mathbf{Z}_{m})$, $\boldsymbol{\gamma}^{(\pi)}_{m}$ and $\boldsymbol{\gamma}^{(A)}_{m, 1:K}$, Infer $\boldsymbol{\gamma}^{(B)}_{m, 1:K}$ as \equationname~\ref{Equation:Update-B-pf}.

\end{document}